\documentclass[letterpaper]{article} 
\def\AAAIFinalVersion{}
\ifdefined\AAAIFinalVersion
  \usepackage{aaai2027} 
\else
  \usepackage[submission]{aaai2027} 
\fi
\usepackage[hyphens]{url} 
\usepackage{graphicx} 
\def\UrlFont{\rm} 
\usepackage{natbib} 
\usepackage{caption} 
\usepackage[utf8]{inputenc}
\usepackage{amsmath,amssymb,amsthm}
\usepackage{booktabs}
\usepackage{array}
\usepackage{tabularx}
\usepackage{colortbl}
\usepackage{verbatim}
\usepackage{multicol}
\graphicspath{{figures/}}
\usepackage{tikz}
\usetikzlibrary{arrows.meta,positioning,fit,backgrounds,calc,decorations.pathmorphing,patterns}
\usepackage[breakable]{tcolorbox}

\definecolor{accent}{HTML}{1F4E79}
\definecolor{softfill}{HTML}{EAF2F8}
\definecolor{validationcolor}{HTML}{00838F}
\definecolor{catalogcolor}{HTML}{1F4E79}
\definecolor{simulatorcolor}{HTML}{2E7D32}
\definecolor{taskcolor}{HTML}{8A5A00}
\definecolor{taskpromptrule}{HTML}{E2E8F0}
\definecolor{taskpromptink}{HTML}{334155}
\definecolor{taskpromptlabel}{HTML}{475569}
\definecolor{taskprompttemplate}{HTML}{B87333}
\definecolor{taskpromptcontext}{HTML}{1F4E79}
\definecolor{taskpromptgold}{HTML}{22C55E}
\definecolor{taskpromptgoldstroke}{HTML}{16A34A}
\definecolor{taskpromptgoldbg}{HTML}{DCFCE7}
\definecolor{taskpromptdistractor}{HTML}{EF4444}
\definecolor{taskpromptdistractorstroke}{HTML}{B91C1C}
\definecolor{taskpromptdistractorbg}{HTML}{FEE2E2}
\definecolor{prompttitle}{HTML}{3F3F3F}
\definecolor{promptborder}{HTML}{4A4A4A}
\definecolor{promptfill}{HTML}{F2F2F2}
\theoremstyle{definition}
\newtheorem{definition}{Definition}

\newcounter{algorithm}
\renewcommand{\thealgorithm}{\arabic{algorithm}}
\newcounter{algorithmicline}
\newcounter{algorithmicindentlevel}
\newlength{\algorithmiclabelwidth}
\newlength{\algorithmiclabelsep}
\newlength{\algorithmicindent}
\newenvironment{algorithm}[1][]{%
  \begin{figure}[t]%
  \refstepcounter{algorithm}%
  \begingroup
  \renewcommand{\caption}[1]{%
    \par\smallskip\noindent\textbf{Algorithm \thealgorithm: }##1\par\smallskip
  }%
  \centering
  \begin{minipage}{0.98\linewidth}
  \footnotesize
}{%
  \end{minipage}%
  \endgroup
  \end{figure}
}
\newenvironment{algorithmic}[1][]{%
  \setcounter{algorithmicline}{0}%
  \setcounter{algorithmicindentlevel}{0}%
  \begin{list}{}{%
    \setlength{\leftmargin}{0pt}%
    \setlength{\labelwidth}{0pt}%
    \setlength{\labelsep}{0pt}%
    \setlength{\itemsep}{0.1em}%
    \setlength{\parsep}{0pt}%
  }%
}{%
  \end{list}
}
\newcommand{\AlgorithmLine}{%
  \stepcounter{algorithmicline}\item[]%
  \parshape=2
    0pt \linewidth
    \dimexpr\algorithmiclabelwidth+\algorithmiclabelsep+\value{algorithmicindentlevel}\algorithmicindent\relax
    \dimexpr\linewidth-\algorithmiclabelwidth-\algorithmiclabelsep-\value{algorithmicindentlevel}\algorithmicindent\relax
  \noindent\makebox[\algorithmiclabelwidth][r]{\footnotesize\thealgorithmicline:}%
  \hspace*{\algorithmiclabelsep}%
  \hspace*{\dimexpr\value{algorithmicindentlevel}\algorithmicindent\relax}%
  \ignorespaces
}
\newcommand{\Require}[1]{\AlgorithmLine \textbf{Input:} #1}
\newcommand{\State}{\AlgorithmLine}
\newcommand{\ForAll}[1]{\AlgorithmLine \textbf{for all} #1 \textbf{do}\stepcounter{algorithmicindentlevel}}
\newcommand{\EndFor}{\addtocounter{algorithmicindentlevel}{-1}\AlgorithmLine \textbf{end for}}
\newcommand{\If}[1]{\AlgorithmLine \textbf{if} #1 \textbf{then}\stepcounter{algorithmicindentlevel}}
\newcommand{\EndIf}{\addtocounter{algorithmicindentlevel}{-1}\AlgorithmLine \textbf{end if}}

\newcommand{\M}{\mathcal{M}}
\newcommand{\Iset}{\mathcal{I}}
\newcommand{\SigmaSet}{\Sigma}
\newcommand{\AlgorithmIndent}{%
  \ifcase\value{algorithmicindentlevel}0em\or1.2em\or2.4em\or3.6em\else4.8em\fi
}
\newcommand{\Statex}{\par\noindent\hspace*{2.2em}}
\newenvironment{promptbox}[1][]{%
  \begin{tcolorbox}[
    breakable,
    use color stack=true,
    colback=promptfill,
    colframe=promptborder,
    colbacktitle=prompttitle,
    coltitle=white,
    colupper=black,
    title={#1},
    adjusted title after break={#1: Continued},
    fonttitle=\bfseries,
    boxrule=0.8pt,
    toprule at break=0.8pt,
    bottomrule at break=0.8pt,
    arc=4pt,
    outer arc=4pt,
    left=8pt,
    right=8pt,
    top=7pt,
    bottom=7pt,
    extras={%
      colback=promptfill,
      colframe=promptborder,
      colbacktitle=prompttitle,
      coltitle=white,
    },
    extras title after break={%
      colback=promptfill,
      colframe=promptborder,
      colbacktitle=prompttitle,
      coltitle=white,
    },
    before skip=6pt,
    after skip=6pt,
    before upper={%
      \footnotesize
      \raggedright
      \parindent=0pt
      \parskip=0.35em
    },
  ]%
}{%
  \end{tcolorbox}
}
\newcommand{\promptsection}[1]{\par\smallskip\noindent\textbf{#1}\par}

\newcommand{\placeholder}[1]{%
  \texttt{\textless}%
  \begingroup\def\UrlFont{\ttfamily}\url{#1}\endgroup%
  \texttt{\textgreater}%
}
\newcommand{\mechex}[1]{\textit{#1}}
\newenvironment{prompttemplateblock}{\par\noindent\begingroup\color{black}}{\par\endgroup}

\newcommand{\code}[1]{%
  \begingroup\def\UrlFont{\ttfamily}\url{#1}\endgroup%
}
\newenvironment{promptitems}{%
  \begin{list}{$\bullet$}{%
    \leftmargin=1.4em
    \itemsep=0.1em
    \parsep=0pt
    \topsep=0.2em
  }%
}{%
  \end{list}
}
\newenvironment{promptmeta}{%
  \begin{list}{}{%
    \leftmargin=0pt
    \itemsep=0.05em
    \parsep=0pt
    \topsep=0.15em
  }%
}{%
  \end{list}
}
\newcommand{\promptfield}[2]{\item \textbf{#1:} #2}

\newcommand{\Yes}{\mbox{\normalfont\scshape yes}}
\newcommand{\No}{\mbox{\normalfont\scshape no}}

\title{MechReasoner: A Simulator and Benchmark for Mechanistic Reasoning in Qualitative Physics}
\ifdefined\AAAIFinalVersion
  \author{
    Danilo Gusicuma\textsuperscript{1,2}
    \quad
    André Freitas\textsuperscript{1,3}
  }
  \affiliations{
    \textsuperscript{1}Idiap Research Institute, Switzerland \\
    \textsuperscript{2}École Polytechnique Fédérale de Lausanne (EPFL), Switzerland \\
    \textsuperscript{3}Department of Computer Science, University of Manchester, United Kingdom \\
    \texttt{firstname.lastname@idiap.ch}
  }
\else
  \author{Anonymous Submission}
  \affiliations{}
\fi

\begin{document}
\maketitle

\begin{abstract}
This work introduces MechReasoner, a mechanistic qualitative simulator grounded
in confluence-based qualitative physics, together with a benchmark for mechanistic
inference. Current large language models (LLMs) generate fluent mechanistic
descriptions that do not reliably follow from underlying structural and causal
constraints. The benchmark tests whether answers preserve simulator-licensed
ambiguity, quantified claims, episode-graph transition evidence, repairs, and
trace-support judgments. Its 1{,}120 items are generated deterministically from
admissible interpretation sets, component states, scenario restrictions,
confluence constraints, and derivation steps across 18 catalog mechanisms and six
task families. Each mechanism undergoes converter checks of structure
and topology and behavioral checks against quantitative simulations.
GPT-5.5 accuracy decreases as
family-specific mechanistic complexity increases, from 76.1\% in the
lowest-complexity bucket (B1) to 38.0\% in the highest-complexity bucket (B4).
The negative association remains after controls for rendered-prompt and
expected-answer length. These results show that qualitative simulators can
support auditable NLP benchmarks for mechanistic inference.
\end{abstract}

\section{Introduction}

Across domains ranging from physical devices and biological systems to economic models,
understanding the world requires reasoning about its underlying mechanisms. However, while
state-of-the-art LLMs, including models designed for agentic use, can generate
fluent descriptions of devices and processes,
they often fail to accurately predict real-world phenomena.

Accounts of mechanisms describe entities and activities organized to produce
changes \cite{machamer2000}. Mechanistic reasoning identifies setup conditions,
entities, their activities and properties, and their organization, then chains
them to explain how a phenomenon arises \cite{russ2008}. For this work, this
means deriving claims about a system's states and changes from an explicit
representation of its mechanism.

Qualitative physics provides such a representation when exact numerical
parameters are unavailable
\cite{dekleer1984,forbus1984,kuipers1984,kuipers1986}. It captures the
structural and causal constraints that govern how a mechanism can behave and
preserves alternative behaviors when the available information is incomplete.
The qualitative simulator gives this representation executable semantics by
enumerating admissible interpretations, reachable episode paths, and the
evidence that licenses them. MechReasoner then assesses whether a reasoner
derives claims supported by those semantic objects, including whether a state
or episode path is possible, whether an effect or repair is necessary, and whether
a derivation is supported.

Current benchmarks rarely score the relation between a mechanism and a claim.
Process and physical commonsense tasks infer answers from text
\cite{dalvi2018propara,tandon2019wiqa,bisk2020piqa}. Recent physics benchmarks
emphasize question answering, with some also providing step-level scoring or
executable solution artifacts
\cite{wang2023newton,zhang2025physreason,imani2026sympybench}. Tests that treat
an LLM as a simulator score one subsequent state or state difference
\cite{wang2024textsimulator}. A correct label can still accompany an impossible
transition, an omitted valid alternative, or evidence unsupported by the stated
laws. Evaluating mechanistic reasoning therefore requires a semantic reference
that enumerates admissible states and histories, distinguishes existential from
universal claims, retains transition and derivation evidence, and varies
complexity while preserving the meaning of correctness.

Two questions organize the study. \emph{RQ1. How accurately do LLMs, including
agentic-capable reasoning models, solve simulator-grounded tasks requiring
mechanistic reasoning?}
\emph{RQ2. How does their performance vary across levels of mechanistic
complexity?}

MechReasoner addresses this gap with a benchmark that evaluates mechanistic
reasoning against an executable qualitative reference.

The benchmark's essential abstraction represents each mechanism as a
constrained possibility space. Components, topology, local laws, operating
conditions, and transition rules define the space. Six task families apply
operations to this shared object. They test state membership, episode path
existence, effect necessity, reachable change projection, repair universality,
and derivation support. The
simulator materializes the space as admissible interpretations, an episode
graph, and auditable evidence.

\begin{figure*}[!t]
\centering
\includegraphics[width=\textwidth]{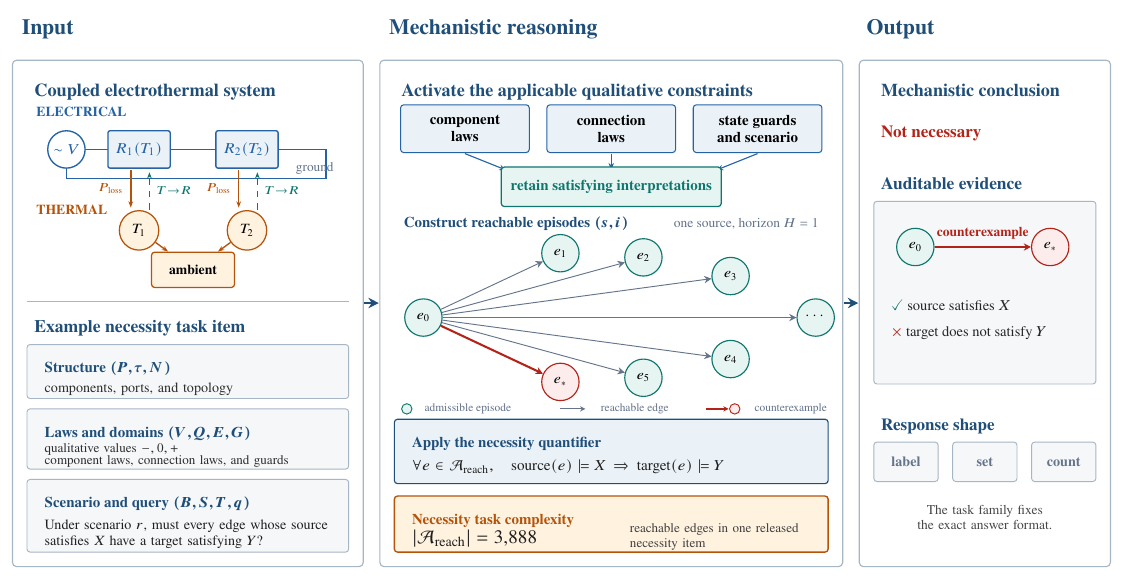}
\caption{Mechanistic reasoning in one necessity task. The input specifies a
coupled electrothermal mechanism, its qualitative structure and laws, a
scenario, and a universal query. Applicable constraints yield admissible
interpretations and a one-step episode graph. The red edge is a counterexample
because its source satisfies $X$ and its target fails to satisfy $Y$.
Necessity complexity counts distinct accepted causal edges reachable within
the task horizon, here
$\lvert\mathcal A_{\mathrm{reach}}\rvert=3{,}888$.}
\label{fig:necessity-mechanistic-reasoning}
\end{figure*}
\FloatBarrier

A qualitative simulator grounded in confluences compiles each encoded
mechanism into its admissible interpretations, episode graph, and audit
evidence. The simulator generates and scores tasks deterministically. Models
receive the persisted text prompts and have no access to the simulator.

Nine model endpoints answer the same 1{,}120 prompts. Their underlying models
span general LLMs and models with dedicated reasoning, with and without
documented agentic readiness, but the evaluated configurations are all direct
LLM inference: tools and multi-step action loops are disabled. Accuracy
requires exact agreement with simulator gold after normalization defined for
each task family.
Simulator complexity bins are formed separately within each task family. Every
model scores lower in the highest complexity bucket than in the lowest after the
task families are pooled. GPT-5.5 declines from 76.1\% to 38.0\%.

Three elements define the benchmark release.
\begin{enumerate}
\item The evaluator contracts express mechanistic reasoning as quantified
queries over finite qualitative states and histories while retaining ambiguity.
\item The dataset contains 1{,}120 structured items across 18 mechanisms
and six task families. The task semantics make existential claims, universal
claims, enumeration of sets, and counts explicit. Gold evidence is inspectable,
and complexity measures are defined within each task family.
\item The release includes the typed representation, qualitative simulator,
deterministic generators, manifests, evaluators, and analysis artifacts needed
to trace each answer to the encoded mechanism. Bounded converter checks and an
independent quantitative consistency check grounded on a mechanistic
simulator cover 341 observed
scenarios across all 18 catalog mechanisms.
\end{enumerate}

The code, catalog, benchmark, and derived artifacts are distributed under
GPL-3.0-only.

\section{Mechanistic Abstraction and Benchmark}

\subsection{Formal Problem Statement}
\label{sec:formal-problem}

For each composite state and scenario, the intrastate solver returns the
admissible interpretations licensed by the active domains and qualitative
constraints. A separate simulation session collects the active transition
declarations and combines them with the solved interpretations to construct
the episode graph. Benchmark labels are derived from these solved objects.

\begin{definition}[Mechanism and interpretations]
A qualitative mechanism is
\[
\M = (P,\tau,V,Q,E,N,S,G,T,B).
\]
Here $P$ is the finite set of component instances, $\tau$ maps each component to
its component type, $V$ is the finite typed variable set attached to components
or boundary inputs, and $Q$ assigns each $v\in V$ a finite qualitative quantity
space $Q(v)$.
$E$ contains component-local qualitative constraints, including
state-dependent confluences, while $N$ contains topology and
connection-law constraints between components. $S$ is the composite
component-state space induced by $P$ and $\tau$, $G$ contains state-dependent
domain restrictions, $T$ contains component-local transition declarations and
their conditions, and $B$ contains admissible scenarios. For a composite
state $s \in S$ and scenario $r \in B$, the compiler instantiates the active
component confluences and connection laws and restricts the variable domains
using the applicable state and scenario restrictions. The intrastate solver
returns
\[
\Iset_{s,r} = \SigmaSet(\M,s,r),
\]
the finite set of interpretations, i.e., the qualitative assignments that
satisfy those active domains and constraints. The simulation session separately
collects $T(s)$ and evaluates those declarations when constructing
episode-graph edges.
\end{definition}

Figure~\ref{fig:necessity-mechanistic-reasoning} follows one necessity item
from its serialized mechanism and query through admissible interpretations and
reachable edges to a conclusion supported by a counterexample.

The simulator preserves each $\Iset_{s,r}$ as a set as a
semantic choice. Within one composite state, a claim is plausible at the
interpretation level when at least one admissible interpretation supports it
and necessary at that level when every admissible interpretation supports it.
These local notions preserve underdetermination across interpretations. The
Necessity task family quantifies over reachable edges with sources that match
the claim (Table~\ref{tab:task-semantics}).

\begin{definition}[Episode graph]
For a scenario $r$, the episode graph
\[
\mathcal G_r = (\mathcal V_r,\mathcal A_r)
\]
contains one node $(s,i)$ for each state $s \in S$ and each interpretation
$i \in \Iset_{s,r}$. An edge in $\mathcal A_r$ connects two distinct nodes when a
simulator-derived cause and the causal, continuity, and endpoint checks license
the qualitative change. Every changed component state must be licensed by a
compatible declaration whose guards hold. An unchanged component needs no
declaration, so the graph may also contain interpretation-only edges.
\end{definition}

These solved objects define the six task families summarized in
Table~\ref{tab:task-semantics}. The supplementary material gives the full
clauses for state restrictions, outgoing-transition witnesses, textual
verification, and repair candidates.

\subsection{Simulator Architecture}
\label{sec:simulator-architecture}

The intrastate solver instantiates component and connection laws, solves every
composite state, and retains its admissible assignments, forced facts, and
evidence. A separate session constructs episode-graph edges by applying
declaration, guard, causal, continuity, and endpoint checks. These records ground
task labels and audits; complexity is computed deterministically from the
mechanism specification and simulator records.

\begin{algorithm}
\caption{Deterministic generation of benchmark tasks}
\label{alg:task-generation}
\begin{algorithmic}[1]
\Require mechanisms $\mathcal M$, task families $\mathcal F$, cell quota $k$,
and prompt-token ceiling $L$
\State $\mathcal R\gets[\,]$
\ForAll{$M\in\mathcal M$}
  \State $Z_M\gets\operatorname{SemanticObjects}(M)$
  \ForAll{$f\in\mathcal F$ supported by $M$}
    \State $\mathcal A_{M,f}\gets[\,]$
    \ForAll{$c\in\operatorname{Candidates}_f(M,Z_M)$}
      \State $y_c\gets\operatorname{Query}_f(c,Z_M)$
      \State $w_c\gets\operatorname{Complexity}_f(c,Z_M)$
      \State $p_c\gets\operatorname{Render}_f(M,c)$
      \If{$\operatorname{Valid}_f(p_c,y_c)$ and
      $\operatorname{Tokens}(p_c)\leq L$}
        \State append $(p_c,y_c,w_c,\operatorname{Audit}(c,Z_M))$ to
        $\mathcal A_{M,f}$
      \EndIf
    \EndFor
    \State $\mathcal R\gets\mathcal R\cup
    \operatorname{SelectLevels}_f(\mathcal A_{M,f},\{\mathrm{D1},\ldots,
    \mathrm{D4}\},k)$
  \EndFor
\EndFor
\State $\mathcal R\gets\operatorname{DeduplicateByNormalizedPrompt}(\mathcal R)$
\State persist tasks, gold solutions, manifests, and completion hashes
\State \textbf{return} $\mathcal R$
\end{algorithmic}
\end{algorithm}

Algorithm~\ref{alg:task-generation} separates the shared simulator records from
the clauses that vary across task families. Table~\ref{tab:task-semantics}
defines each query, Table~\ref{tab:family-complexity-metrics} defines each
complexity value, and the supplement specifies candidate construction,
validation, and rendering.

\definecolor{neutralstroke}{HTML}{64748B}
\definecolor{neutralbg}{HTML}{EEF2F7}
\begin{table*}[!t]
\centering
\tikzset{
  vb/.style={circle,draw=taskpromptgoldstroke,fill=taskpromptgoldbg,
             inner sep=1pt,minimum size=9mm,font=\small,align=center},
  xb/.style={circle,draw=taskpromptdistractorstroke,fill=taskpromptdistractorbg,
             inner sep=1pt,minimum size=9mm,font=\small,align=center},
  nb/.style={circle,draw=neutralstroke,fill=neutralbg,
             inner sep=1pt,minimum size=9mm,font=\small,align=center},
  ed/.style={-{Stealth[length=4pt]},neutralstroke,line width=0.6pt},
  edg/.style={-{Stealth[length=4pt]},taskpromptgoldstroke,line width=0.7pt},
  edr/.style={-{Stealth[length=4pt]},taskpromptdistractorstroke,line width=0.7pt,dashed},
  fl/.style={font=\small\itshape,text=taskpromptlabel},
  tg/.style={font=\small\bfseries},
  num/.style={font=\small,text=taskpromptlabel},
}
\newcommand{\xmark}[1]{\draw[taskpromptdistractorstroke,line width=0.8pt,line cap=round]
  (#1.south west)--(#1.north east) (#1.north west)--(#1.south east);}
\newcommand{\visualmargin}{\path[use as bounding box]
  ([yshift=-3pt]current bounding box.south west) rectangle
  ([yshift=3pt]current bounding box.north east);}
\setlength{\tabcolsep}{6pt}
\renewcommand{\arraystretch}{1.35}
\small
\begin{tabularx}{\textwidth}{@{}>{\raggedright\arraybackslash}p{3.0cm} >{\raggedright\arraybackslash}X r@{}}
\toprule
Task family & Decision criterion & Illustrative figure \\
\midrule
State consistency
& A candidate composite state is consistent iff there exists a complete
qualitative interpretation satisfying the selected component states, operating
context, case restrictions, and all active laws.
&
\begin{tikzpicture}[baseline=0pt]
  \node[vb] (s1) at (0,0){$s_1$};
  \node[xb] (s2) at (1.1,0){$s_2$};
  \node[vb] (s3) at (2.2,0){$s_3$};
  \node[xb] (s4) at (3.3,0){$s_4$};
  \node[vb] (s5) at (4.4,0){$s_5$};
  \xmark{s2}\xmark{s4}
  \visualmargin
\end{tikzpicture}
\\
\midrule
Plausibility
& A narrative is plausible (P) iff there exists an $H$-step path from the
initial episode such that the episode at every checkpoint satisfies all
observations listed for that checkpoint; otherwise it is implausible (I).
&
\begin{tikzpicture}[baseline=0pt]
  \node[nb] (e0) at (0,0){$e_0$};
  \node[vb] (e1) at (1.2,0){$e_1$};
  \node[vb] (e2) at (2.4,0){$e_2$};
  \node[num,above=0.5pt of e0]{$t=0$};
  \node[num,above=0.5pt of e1]{$O_1$};
  \node[num,above=0.5pt of e2]{$O_2$};
  \draw[edg] (e0)--(e1);
  \draw[edg] (e1)--(e2);
  \node at (3.15,0){$\Rightarrow$};
  \node[font=\small\bfseries,anchor=west] at (3.45,0){P};
  \visualmargin
\end{tikzpicture}
\\
\midrule
Necessity
& Every retained claim has at least one supplied reachable edge whose source
satisfies all SOURCE facts. A claim is necessary (N) iff every such edge has a
target satisfying at least one TARGET alternative; otherwise it is not
necessary (U).
&
\begin{tikzpicture}[baseline=0pt]
  \node[vb] (r1) at (0,0){$r_1$};
  \node[vb] (r2) at (1.1,0){$r_2$};
  \node[vb] (r3) at (2.2,0){$r_3$};
  \node at (3.05,0){$\Rightarrow$};
  \node[font=\small,align=left,text=black,anchor=west] at (3.35,0)
    {$\#\mathrm{CE}=0:\ \mathbf{N}$};
  \visualmargin
\end{tikzpicture}
\\
\midrule
Episode-graph transitions
& From the reachable edges up to horizon $H$, list every distinct component
state change as a (component, source state, target state) triple. Exclude
declared changes not realized by an edge and edges that change only the
qualitative interpretation.
&
\begin{tikzpicture}[baseline=0pt]
  \node[nb] (q0) at (0,0){$q_0$};
  \node[vb] (q1) at (1.4,0){$q_1$};
  \node[nb] (q2) at (2.8,0){$q_2$};
  \node[xb] (q3) at (4.2,0){$q_3$};
  \draw[edg] (q0)--(q1);
  \draw[edr] (q2)--(q3);
  \visualmargin
\end{tikzpicture}
\\
\midrule
Functional recovery
& Classify each policy over all named post-repair outcomes and their one-step
successors: unsafe (U) if any is unsafe; otherwise deadline failure (D) if any
deadline episode fails the goal; otherwise recovered (R).
&
\begin{tikzpicture}[baseline=0pt]
  \node[nb] (policy) at (0,0){$\pi$};
  \node[nb] (outcome) at (2.2,0){$o$};
  \node[xb] (successor) at (4.4,0){$o'$};
  \draw[ed] (policy)-- node[fl,above,inner sep=1pt]{names} (outcome);
  \draw[edr] (outcome)-- node[fl,above,inner sep=1pt]{one step} (successor);
  \node[font=\small,align=center,text=black] at (0,-0.82) {policy};
  \node[font=\small,align=center,text=black] at (2.2,-0.82) {outcome};
  \node[font=\small,align=center,text=black] at (4.4,-0.82)
    {unsafe $\Rightarrow$ U};
  \visualmargin
\end{tikzpicture}
\\
\midrule
Trace faithfulness
& Replay each trace in order. A step is unfaithful if its displayed before-set
differs from the current set or if its cited law does not produce exactly its
displayed after-set. Continue from the displayed after-set even after an
unfaithful step, and count all such steps.
&
\begin{tikzpicture}[baseline=0pt]
  \node[nb] (f1) at (0,0){$f_1$};
  \node[nb] (f2) at (1.2,0){$f_2$};
  \node[xb] (f3) at (2.4,0){$f_3$};
  \node[nb] (f4) at (3.6,0){$f_4$};
  \node[num,above=0.5pt of f1]{step 1};
  \node[num,above=0.5pt of f2]{step 2};
  \node[num,above=0.5pt of f3]{step 3};
  \node[num,above=0.5pt of f4]{step 4};
  \draw[ed] (f1)--(f2);
  \draw[edr] (f2)--(f3);
  \draw[ed] (f3)--(f4);
  \visualmargin
\end{tikzpicture}
\\
\bottomrule
\end{tabularx}
\caption{Task families and the problem semantics each family tests, shown as the
illustrative figure behind each task. Green marks valid or forced elements,
red marks invalid, blocked, or unsupported ones, and neutral circles are candidates
under test. In the illustrative figures, $s_i$ denotes complete candidate
component-state vectors, $e_i$ path episodes, $O_t$ checkpoint observations,
$r_i$ reachable edge rows, $q_i$ component states, $\pi$ a repair policy,
$o,o'$ a named outcome and successor, and $f_i$ trace steps.}
\label{tab:task-semantics}
\end{table*}

\subsubsection{Quantitative--qualitative consistency check}

The profile admits 12 Modelica Standard Library v4.1.0
examples~\cite{modelicastandardlibrary410} and six compositions of supported
components, spanning electrical, translational, rotational, and thermal models.
After freezing the outputs of 72 bounded OpenModelica runs
\cite{openmodelica1269}, the check covers 341 distinct observed scenarios across
all 18 mechanisms. The 341 scenarios comprise 186 projected composite-state
configurations and 155 observed component-transition shapes. The numerical and
qualitative results follow separate execution paths, and the comparison reads
neither benchmark tasks nor gold solutions. The contained
experiment has zero contradictions and zero inconclusive cases. The supplement
details the observations and their acceptance rules.
Modelica Standard Library and OpenModelica remain external
dependencies.

\subsection{Benchmark as a Mechanistic Abstraction}

At the benchmark level, a mechanism is not identified with a particular
simulator implementation or with one predicted trajectory. For scenario $r$,
its semantic abstraction comprises the finite interpretation sets
$\{\Iset_{s,r}\}_{s\in S}$, the episode graph $\mathcal G_r$, and the rule-evidence
records that justify their construction. Each item selects a bounded view of
this object and applies a query $q$ whose answer is fixed by the released
semantics. The simulator constructs the object and computes the gold answer;
the benchmark evaluates whether another reasoner can recover the requested
property from its serialized representation. This separation makes the
benchmark about mechanistic reasoning rather than simulator imitation.

\subsubsection{Task semantics}

The benchmark is organized around mechanistic questions and currently comprises six task families,
each designed to probe a distinct aspect of the problem. Table~\ref{tab:task-semantics} summarizes
 the relationship between these families and the semantics they test.
\textbf{State consistency} identifies complete candidate component-state
vectors admitting a satisfying qualitative assignment; \textbf{plausibility}
labels checkpoint narratives witnessed by one exact bounded episode path;
\textbf{necessity} labels source-to-target claims with no counterexample in a
supplied exact edge graph; \textbf{episode-graph transition tasks} enumerate
distinct component-state changes witnessed by reachable edges;
\textbf{functional recovery} assigns universal one-step safety/deadline statuses
to repair policies; and \textbf{trace faithfulness} counts stale or unsupported
displayed propagation steps.

\begin{table*}[!t]
\centering
\small
\setlength{\tabcolsep}{4pt}
\begin{tabularx}{\textwidth}{>{\raggedright\arraybackslash}p{2.7cm} >{\raggedright\arraybackslash}p{3.9cm} >{\raggedright\arraybackslash}X}
\toprule
Task family & Complexity scalar & What the scalar counts \\
\midrule
State consistency &
$\sum_{i=1}^{n_{\mathrm{cand}}}\sum_{k=1}^{n_{\mathrm{case}}}
(L_i+R_i+A_k)$ &
Active confluences and state and case restrictions for every displayed
candidate--case pair. \\
Plausibility &
$\sum_{t=1}^{H}|O_t|$ &
Displayed qualitative facts across the successor checkpoints in one candidate
narrative. \\
Necessity &
$|\mathcal A_{\mathrm{reach}}|$ &
Distinct accepted causal edges reachable within the task horizon.\\
Episode-graph transitions &
$\sum_{x\in\mathcal{X}_{\mathrm{exp}}}B_xT_x$ &
Possible boundary-cause and declared-transition pairings at each reachable
source episode. \\
Functional recovery &
$P+\sum_{p=1}^{P}E_p$ &
Displayed repair plans and their distinct accepted one-step effects. \\
Trace faithfulness &
$\sum_{g=1}^{n_{\mathrm{tr}}} h_g$ &
Displayed trace-step occurrences across the independent traces. \\
\bottomrule
\end{tabularx}
\caption{Metrics for mechanistic complexity computed for each task family
without using the expected answer.}
\label{tab:family-complexity-metrics}
\end{table*}

\subsubsection{Benchmark governance}

The audited comparator pool is deduplicated by normalized prompt. Completion
markers bind the catalog, generator, manifest, solutions, and task files. Models
receive only persisted prompts with their expected response shape. All fit an 80{,}000-token
\texttt{o200k\_base} ceiling; the maximum is 73{,}866 tokens.

\subsubsection{Evaluator protocol}

Each backend receives the same fixed system message and persisted prompt,
without tools or repository access. No family schema is sent out of band;
supporting endpoints receive only generic object mode. The scorer extracts
the returned answer, normalizes it by family, and compares it exactly with gold.
Rejected outputs are protocol failures, accepted non-gold outputs semantic
failures, and request or scorer errors infrastructure failures. End-to-end
accuracy retains every model--item record. Family-specific component measures
include label-character and candidate--case accuracy, transition-set F1, and
trace count-field accuracy. The supplement specifies request controls, retries,
message envelopes, normalization, exact evaluator profiles, effective output
limits, and reasoning-effort settings.

\section{Empirical Analysis}

\subsection{Experiment Design}

\subsubsection{Model Selection Criteria}

The nine models span disclosed dense and sparse scales, proprietary and open
weights, dedicated reasoning variants, and general, multilingual, code, STEM,
and agentic training profiles; GPT-5.5 supplies the frontier reference point.
The supplementary \emph{Model Inventory} separately records dedicated
reasoning, documented agentic readiness, and the direct-LLM system class used
in this experiment.

\subsubsection{Mechanistic Complexity}
\label{sec:mechanistic-complexity}

Table~\ref{tab:family-complexity-metrics} lists the scalar for each task family.
Each scalar is a deterministic, answer-independent count derived from the
qualitative simulator. Retained items are re-scored by the active family metric
before complexity buckets and reported slices are computed.

One deterministic B1--B4 analysis-bin assignment is used for every reported
figure, table, and control. Within each task family, deterministic,
tie-preserving near-quartiles keep every item with the same complexity score
in one bin. Each cut is placed at the tie-block boundary nearest its cumulative
equal-count target; an exact distance tie selects the earlier boundary.

The labels identify complexity strata within each task family. They do not make
the absolute complexity scalars comparable across families. Cross-family
comparisons align these strata, and item records retain the concrete quantities
defined for their task family. The analysis bins are distinct from the
generation-time D1--D4 labels stored with each task. Applied to the 1{,}120
retained questions, the assignment gives 309/273/267/271 items in B1/B2/B3/B4.

\paragraph{Accuracy controls}
An exploratory stacked conditional-logistic analysis stacks all 10{,}080
model--item records, with strata formed by model, task family, and mechanism
and CR1 inference clustered by mechanism. The adjusted regression adds log
prompt and expected-answer character counts. The supplement reports the
fully enumerated wild-cluster test and regression diagnostics.

\subsubsection{Dataset Characterization}

Table~\ref{tab:dataset-characterization} summarizes the released benchmark
before model scores are considered. The 108 family--mechanism cells each target
16 structured tasks, giving 1{,}728 targeted
generation slots. Generation produced 1{,}141 task-ID records from 50 complete,
35 partial, and 23 empty pools. Normalized-prompt deduplication removed 21
repeated records, leaving 1{,}120 unique questions across six task families and 18
mechanisms. The resulting 608-question shortfall is not padded. The mechanism
column reports how many mechanisms supply retained items inside each family.

\begin{table}[!b]
\centering
\small
\begin{tabularx}{\linewidth}{>{\raggedright\arraybackslash}X >{\centering\arraybackslash}p{0.95cm} >{\centering\arraybackslash}p{1.05cm}}
\toprule
Task family & Items & Mech. \\
\midrule
State consistency & 212 & 17 \\
Plausibility & 136 & 11 \\
Necessity & 160 & 14 \\
Episode-graph transitions & 100 & 11 \\
Functional recovery & 288 & 18 \\
Trace faithfulness & 224 & 14 \\
\midrule
Total & 1{,}120 & 18 \\
\bottomrule
\end{tabularx}
\caption{Compact characterization of the benchmark.}
\label{tab:dataset-characterization}
\end{table}

\subsection{Results}

\begin{figure*}[!t]
\centering
\resizebox{\textwidth}{!}{%
  \input{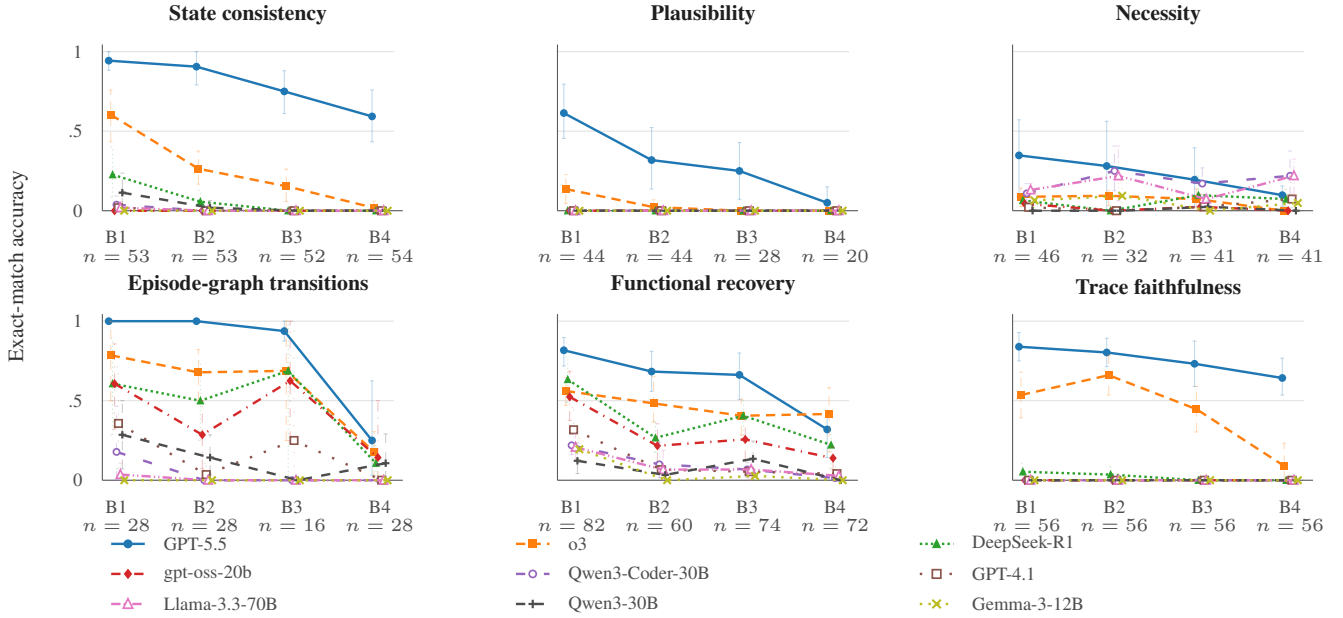}%
}
\caption{Accuracy of all nine direct-inference model endpoints across the canonical bins of the
simulator-complexity proxy defined relative to each task family. The shared
legend applies to every panel, which also shows its bin sample sizes; error
bars are 95\% percentile intervals from 10{,}000 item-weighted
mechanism-level cluster-bootstrap replicates. B1--B4 order complexity only within
each family; the six raw scalars retain different units.}
\label{fig:gpt-family-complexity-curves}
\end{figure*}

\paragraph{Accuracy declines substantially with mechanistic complexity}
Figure~\ref{fig:gpt-family-complexity-curves} compares all nine direct-inference
model endpoints. GPT-5.5,
the strongest model and clearest view above the accuracy floor, has lower B4
than B1 accuracy in all six families, with declines of 19.6--75.0 points.
Across the nine models, pooled accuracy falls from 23.7\% in B1 to 8.7\% in
B4; every model scores lower in B4 than B1. The adjusted stacked regression gives a per-bin
odds ratio of 0.619 (mechanism-cluster CR1--$t_{17}$ 95\% interval
$[0.547,0.701]$). The association is defined within each family and does not
support a causal interpretation.

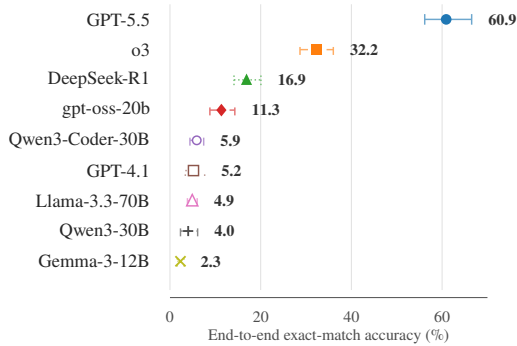
\begin{figure}[!t]
\centering
\begin{tikzpicture}[x=1cm,y=1cm,
  gridline/.style={draw=black!12,line width=0.3pt},
  axis/.style={draw=black!65,line width=0.4pt},
  modellabel/.style={font=\scriptsize,text=black!85},
  ticklabel/.style={font=\tiny,text=black!70},
  valuelabel/.style={font=\tiny\bfseries,text=black!80}
]
\definecolor{plotseries0}{HTML}{1F77B4}
\definecolor{plotseries1}{HTML}{FF7F0E}
\definecolor{plotseries2}{HTML}{2CA02C}
\definecolor{plotseries3}{HTML}{D62728}
\definecolor{plotseries4}{HTML}{9467BD}
\definecolor{plotseries5}{HTML}{8C564B}
\definecolor{plotseries6}{HTML}{E377C2}
\definecolor{plotseries7}{HTML}{4D4D4D}
\definecolor{plotseries8}{HTML}{BCBD22}
\draw[gridline] (0.000,-0.380) -- (0.000,3.400);
\node[ticklabel,anchor=north] at (0.000,-0.530) {0};
\draw[gridline] (1.200,-0.380) -- (1.200,3.400);
\node[ticklabel,anchor=north] at (1.200,-0.530) {20};
\draw[gridline] (2.400,-0.380) -- (2.400,3.400);
\node[ticklabel,anchor=north] at (2.400,-0.530) {40};
\draw[gridline] (3.600,-0.380) -- (3.600,3.400);
\node[ticklabel,anchor=north] at (3.600,-0.530) {60};
\draw[axis] (0,-0.480) -- (4.200,-0.480);
\node[modellabel,anchor=east] at (-0.14,3.200) {GPT-5.5};
\draw[plotseries0!65,solid,line width=0.55pt] (3.370,3.200) -- (3.991,3.200);
\draw[plotseries0!65,solid,line width=0.55pt] (3.370,3.130) -- (3.370,3.270);
\draw[plotseries0!65,solid,line width=0.55pt] (3.991,3.130) -- (3.991,3.270);
\fill[plotseries0] (3.654,3.200) circle (2.00pt);
\node[valuelabel,anchor=west] at (4.091,3.200) {60.9};
\node[modellabel,anchor=east] at (-0.14,2.800) {o3};
\draw[plotseries1!65,densely dashed,line width=0.55pt] (1.720,2.800) -- (2.161,2.800);
\draw[plotseries1!65,densely dashed,line width=0.55pt] (1.720,2.730) -- (1.720,2.870);
\draw[plotseries1!65,densely dashed,line width=0.55pt] (2.161,2.730) -- (2.161,2.870);
\fill[plotseries1] (1.867,2.733) rectangle (2.001,2.867);
\node[valuelabel,anchor=west] at (2.261,2.800) {32.2};
\node[modellabel,anchor=east] at (-0.14,2.400) {DeepSeek-R1};
\draw[plotseries2!65,densely dotted,line width=0.55pt] (0.847,2.400) -- (1.203,2.400);
\draw[plotseries2!65,densely dotted,line width=0.55pt] (0.847,2.330) -- (0.847,2.470);
\draw[plotseries2!65,densely dotted,line width=0.55pt] (1.203,2.330) -- (1.203,2.470);
\fill[plotseries2] (1.013,2.484) -- (0.939,2.333) -- (1.086,2.333) -- cycle;
\node[valuelabel,anchor=west] at (1.303,2.400) {16.9};
\node[modellabel,anchor=east] at (-0.14,2.000) {gpt-oss-20b};
\draw[plotseries3!65,dash dot,line width=0.55pt] (0.528,2.000) -- (0.858,2.000);
\draw[plotseries3!65,dash dot,line width=0.55pt] (0.528,1.930) -- (0.528,2.070);
\draw[plotseries3!65,dash dot,line width=0.55pt] (0.858,1.930) -- (0.858,2.070);
\fill[plotseries3] (0.680,2.084) -- (0.748,2.000) -- (0.680,1.916) -- (0.613,2.000) -- cycle;
\node[valuelabel,anchor=west] at (0.958,2.000) {11.3};
\node[modellabel,anchor=east] at (-0.14,1.600) {Qwen3-Coder-30B};
\draw[plotseries4!65,loosely dashed,line width=0.55pt] (0.264,1.600) -- (0.446,1.600);
\draw[plotseries4!65,loosely dashed,line width=0.55pt] (0.264,1.530) -- (0.264,1.670);
\draw[plotseries4!65,loosely dashed,line width=0.55pt] (0.446,1.530) -- (0.446,1.670);
\draw[plotseries4,line width=0.55pt,fill=white] (0.354,1.600) circle (0.056);
\node[valuelabel,anchor=west] at (0.546,1.600) {5.9};
\node[modellabel,anchor=east] at (-0.14,1.200) {GPT-4.1};
\draw[plotseries5!65,loosely dotted,line width=0.55pt] (0.203,1.200) -- (0.459,1.200);
\draw[plotseries5!65,loosely dotted,line width=0.55pt] (0.203,1.130) -- (0.203,1.270);
\draw[plotseries5!65,loosely dotted,line width=0.55pt] (0.459,1.130) -- (0.459,1.270);
\draw[plotseries5,line width=0.55pt,fill=white] (0.244,1.133) rectangle (0.378,1.267);
\node[valuelabel,anchor=west] at (0.559,1.200) {5.2};
\node[modellabel,anchor=east] at (-0.14,0.800) {Llama-3.3-70B};
\draw[plotseries6!65,densely dash dot dot,line width=0.55pt] (0.231,0.800) -- (0.360,0.800);
\draw[plotseries6!65,densely dash dot dot,line width=0.55pt] (0.231,0.730) -- (0.231,0.870);
\draw[plotseries6!65,densely dash dot dot,line width=0.55pt] (0.360,0.730) -- (0.360,0.870);
\draw[plotseries6,line width=0.55pt,fill=white] (0.295,0.884) -- (0.221,0.733) -- (0.369,0.733) -- cycle;
\node[valuelabel,anchor=west] at (0.460,0.800) {4.9};
\node[modellabel,anchor=east] at (-0.14,0.400) {Qwen3-30B};
\draw[plotseries7!65,dashed,line width=0.55pt] (0.139,0.400) -- (0.366,0.400);
\draw[plotseries7!65,dashed,line width=0.55pt] (0.139,0.330) -- (0.139,0.470);
\draw[plotseries7!65,dashed,line width=0.55pt] (0.366,0.330) -- (0.366,0.470);
\draw[plotseries7,line width=0.70pt] (0.174,0.400) -- (0.308,0.400) (0.241,0.333) -- (0.241,0.467);
\node[valuelabel,anchor=west] at (0.466,0.400) {4.0};
\node[modellabel,anchor=east] at (-0.14,0.000) {Gemma-3-12B};
\draw[plotseries8!65,dotted,line width=0.55pt] (0.097,0.000) -- (0.186,0.000);
\draw[plotseries8!65,dotted,line width=0.55pt] (0.097,-0.070) -- (0.097,0.070);
\draw[plotseries8!65,dotted,line width=0.55pt] (0.186,-0.070) -- (0.186,0.070);
\draw[plotseries8,line width=0.70pt] (0.072,-0.067) -- (0.206,0.067) (0.072,0.067) -- (0.206,-0.067);
\node[valuelabel,anchor=west] at (0.286,0.000) {2.3};
\node[ticklabel,anchor=north] at (2.100,-0.780) {End-to-end exact-match accuracy (\%)};
\end{tikzpicture}
\caption{Overall end-to-end accuracy on the
common 1{,}120-question benchmark. Numbers label point estimates; horizontal bars show 95\% percentile
intervals from 10{,}000 item-weighted mechanism-level cluster-bootstrap
replicates.}
\label{fig:overall-accuracy-by-model}
\end{figure}

\paragraph{Frontier and reasoning models lead the full benchmark}
Figure~\ref{fig:overall-accuracy-by-model} compares overall end-to-end
accuracy. GPT-5.5 reaches 60.9\%, followed by o3 at 32.2\% and
DeepSeek-R1 at 16.9\%. Notably, gpt-oss-20b at low reasoning effort
reaches 11.3\%, followed by Qwen3-Coder-30B at 5.9\% and GPT-4.1 at
5.2\%.
All models are evaluated on the same 1{,}120 items.

\paragraph{Accuracy and component-level performance differ by task family}
By accuracy, GPT-5.5 is strongest on state
consistency (79.7\%), followed by
episode-graph transitions (78.0\%), trace faithfulness (75.4\%), functional
recovery (62.5\%), plausibility (36.0\%), and necessity (23.1\%). Functional recovery
remains comparatively strong after the prompt hides
the simulator counts that determine each plan verdict, separating constraint-
based recovery inference from the other mechanistic skills under one evaluator
and catalog.
The corresponding component scores are 98.4\% candidate--case accuracy for
state consistency, 89.2\% plausibility character accuracy, 65.6\% necessity
character accuracy, 88.3\% transition-set F1, 87.3\% functional-recovery
character accuracy, and 75.4\% trace count-field accuracy. The gap shows that a
structured answer that misses the accuracy criterion often contains many
correct components. Accuracy counts an answer as correct only when its
canonical family payload,
after the documented extraction and normalization rules, equals gold; component
measures capture recovered labels or set elements. Full
per-model, per-family results and constant baselines are in the supplement. The
non-oracle prompt-text symbolic baseline reconstructs each typed task from the
rendered prompt and executes the public qualitative semantics, solving all
1{,}120 items with 100\% accuracy.

\section{Related Work}

\noindent \textbf{Classical qualitative reasoning and simulation.}
MechReasoner follows qualitative physics, where system behavior emerges from
structural models expressed through qualitative variables and confluences.

\vspace{6pt plus 2pt minus 2pt}\noindent
De Kleer and Brown's ENVISION architecture separates device structure, qualitative
variables, qualitative calculus, and connection laws. Its runtime separates
intrastate constraint solving from interstate changes in component state
\cite{dekleer1984}. Component libraries, topology, inputs, boundary conditions,
and active confluences determine allowable states, variable behaviors,
transitions, and explanations. MechReasoner retains the full interpretation set
to support quantified claims, constrained repair analysis, and explanation
checks. This representation aligns with Forbus's process-centered abstraction
\cite{forbus1984} and Kuipers's behavior-from-structure reasoning and qualitative
simulation procedure \cite{kuipers1984,kuipers1986}. Falkenhainer and Forbus
show how larger physical domain theories can be assembled compositionally for
task-specific reasoning \cite{falkenhainer1991}.

\noindent Klenk et al.\ provide the closest technical predecessor. They give a
sound and effective mapping from Modelica models to qualitative reasoning,
extend envisioning to support Modelica's declarative events, and infer three
classes of constraints that reduce unrealizable qualitative trajectories
\cite{klenk2014modelica}. Their objective is qualitative analysis of engineering
models during design. MechReasoner uses the released qualitative semantics and
simulator as the executable grounding for a benchmark.

\noindent \textbf{LLM-era simulator grounding and evaluation.}
Mind's Eye converts a physical question to MuJoCo code and injects the simulated
outcome as a hint for the answering LM. Utopia measures answers based on
relative comparisons across 39 tasks in six physical scenes
\cite{liu2022mindseye}. MechReasoner uses its simulator to construct and score
the benchmark, while evaluated models receive the qualitative mechanism without
simulator outcomes. Gold retains the entire ambiguity set of admissible
interpretations, so the evaluator can distinguish possible from necessary
conclusions under physical underdetermination.

\smallskip\noindent Wang et al.\ provide the closest LLM-as-simulator evaluation. Their
LLM-Sim contract predicts one subsequent text game state, its state difference,
or game progress, with separate measurements for action-driven and
environment-driven transitions \cite{wang2024textsimulator}. In contrast,
MechReasoner evaluates quantified queries over a documented space of admissible
states and histories rather than a single predicted successor.

\noindent \textbf{Process and physical commonsense benchmarks from language.}
Later NLP benchmarks probe related reasoning abilities from text rather than
from executable mechanism models. ProPara evaluates state tracking in process
descriptions, WIQA asks perturbation and influence questions over procedural
text, and PIQA tests physical commonsense over natural-language goals and
candidate actions \cite{dalvi2018propara,tandon2019wiqa,bisk2020piqa}. These
resources provide realistic linguistic coverage and human-authored benchmark
settings. They leave reusable device structure and simulator-grounded
admissible interpretation sets outside the reported evidence. The benchmark here
is narrower and more mechanistic because its task instances expose the objects
that separate plausibility from necessity.

\noindent \textbf{Recent physics benchmarks.} Recent
LLM-era benchmarks scale physics-oriented testing with broader benchmark
pipelines. NEWTON introduces an object-attribute repository and templated
question generation for physical reasoning \cite{wang2023newton}; PhysReason
focuses on multi-step physics problems with answer-level and step-level scoring
\cite{zhang2025physreason}; and SymPyBench provides dynamic, parameterized
physics tasks with executable solution artifacts for controlled auditing
\cite{imani2026sympybench}. These benchmarks measure physical and scientific
reasoning via question answering, whereas this one evaluates against executable
mechanism models whose state and transition semantics are exposed to the evaluator.

\section{Conclusion}

Mechanistic reasoning requires judging claims against the mechanisms that
license them. MechReasoner makes that relation testable by using
qualitative physics to construct a constrained possibility space. Claims follow
from admissible interpretations or reachable episode paths, while diverse
queries preserve underdetermination and support quantified judgments over the
same encoded mechanism.

The prompt-text symbolic baseline solves every item, confirming that the prompts
encode the released semantics. Direct-inference model accuracy declines as
mechanistic complexity increases, and component scores above exact scores show
partial structural recovery can fail to produce a coherent final judgment.
Prompt and answer length controls leave the association intact, indicating that
surface length alone does not account for difficulty with the required
mechanistic operations.

\section{Limitations}

The benchmark is limited to 18 mechanisms and six task families. It does not
measure agentic deployment, interactive tools, multimodal perception,
continuous prediction, open-ended explanation, or safe engineering practice.
The results for agentic-ready models therefore characterize direct inference,
not an agent loop; nor does success establish mechanistic internal computation.
Exact accuracy compounds errors across fields. Prompt- and answer-length
controls leave the complexity association negative and do not remove the
surface-form confound.

Generation coverage is nonuniform across task families and mechanisms.
Bootstrap intervals clustered by mechanism account for shared task structure,
although coverage selection remains; protocol failures stay in the end-to-end
denominator. The reported intervals are conditional on one persisted model
response per item and capture variation from resampling the mechanism catalog.
Variability across repeated model inferences lies outside these intervals.

The prompt-text symbolic baseline shares the released semantics implementation;
its 100\% accuracy establishes prompt completeness and implementation
consistency, not independent gold correctness. The quantitative consistency
check does not ensure full benchmark simulation validation. Simulation-based
validation is limited to behavior observed in finite runs and therefore cannot
validate existential claims for which no supporting behavior was observed. The
qualitative abstraction itself follows established formulations in the
qualitative-reasoning literature
\cite{dekleer1984,forbus1984,kuipers1984,kuipers1986,klenk2014modelica}. Within these limits,
the check adds task- and gold-independent numerical evidence, yielding no contradiction
across all 341 observed scenarios.

\enlargethispage{3\baselineskip}
\section*{Ethical Statement}

Tasks are synthetic and contain no human-subject data. Scores do not establish
real-world engineering competence or certify safety-critical deployment. AI
assistance supported wording review; the authors independently verified all
claims and remain responsible for the submission.

\ifdefined\AAAIFinalVersion
  \section*{Acknowledgements}

  This work was carried out within the Horizon Europe Marie Skłodowska-Curie
  Actions Doctoral Network GenAIDE (Grant Agreement No. 101226927).
  This work has received funding from the Swiss State Secretariat for Education,
  Research and Innovation (SERI).
\fi

\clearpage
\appendix

\definecolor{taskprompttemplate}{HTML}{000000}
\renewenvironment{algorithm}[1][]{%
  \begin{figure}[!t]%
  \refstepcounter{algorithm}%
  \begingroup
  \renewcommand{\caption}[1]{%
    \par\smallskip\noindent\textbf{Algorithm \thealgorithm: }##1\par\smallskip
  }%
  \centering
  \begin{minipage}{0.98\linewidth}
  \footnotesize
}{%
  \end{minipage}%
  \endgroup
  \end{figure}
}
\renewenvironment{algorithmic}[1][]{%
  \setcounter{algorithmicline}{0}%
  \setcounter{algorithmicindentlevel}{0}%
  \par
}{%
  \par
}
\renewcommand{\AlgorithmLine}{%
  \par\stepcounter{algorithmicline}%
  \parshape=2
    0pt \linewidth
    \dimexpr2.2em+\AlgorithmIndent\relax
    \dimexpr\linewidth-2.2em-\AlgorithmIndent\relax
  \noindent\makebox[1.8em][r]{\footnotesize\thealgorithmicline:}%
  \hspace*{0.4em}%
  \hspace*{\AlgorithmIndent}%
  \ignorespaces
}

\twocolumn[{%
  \begin{minipage}[t]{\textwidth}
  \begin{multicols}{2}
  \section{Supplementary Material}

  \subsection{Simulator Architecture Diagram}
  \label{app:simulator-architecture-figure}

  Figure~\ref{fig:architecture} summarizes the artifact and trust boundaries
  from bounded Modelica conversion through benchmark-item persistence.

  \subsection{Accuracy Surface Controls}
  \label{app:exact-match-controls}

  The exploratory robustness analysis stacks all nine evaluated models over the
  common benchmark with 1{,}120 questions and six families, retaining protocol
  failures as incorrect. The resulting 10{,}080 model--item records use the
  canonical B assignment for each task, with 309/273/267/271 questions in
  B1/B2/B3/B4 for each model.
  \end{multicols}

  \centering
  \includegraphics[width=\textwidth]{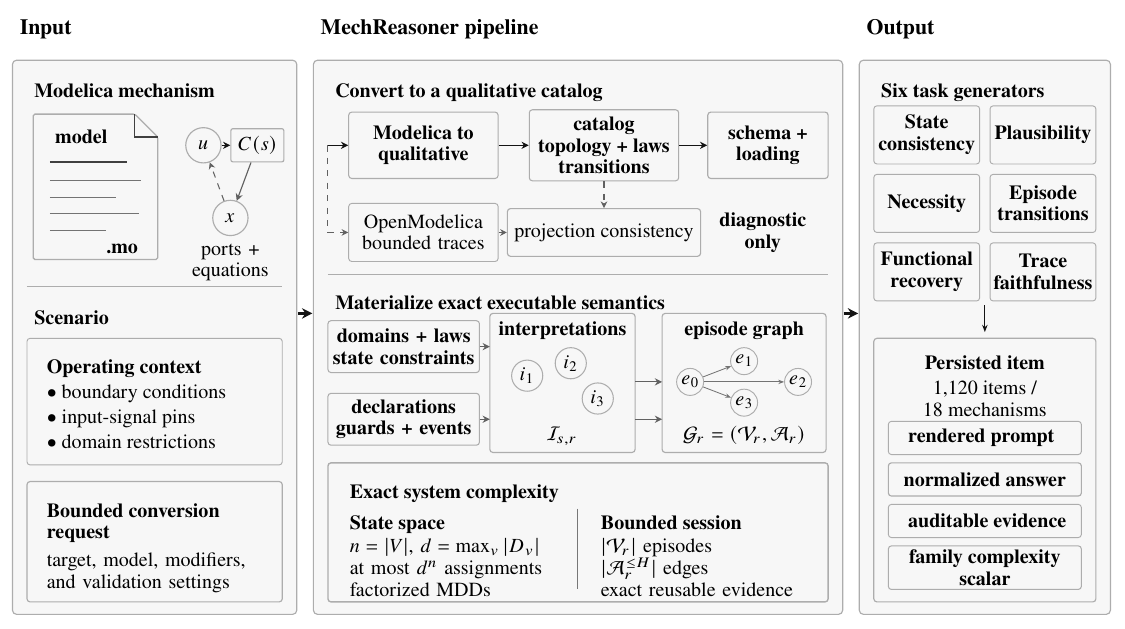}
  \captionof{figure}{MechReasoner system workflow and trust boundaries. Dashed
  gray arrows mark the diagnostic-only OpenModelica branch, which cannot affect
  catalog admission, task generation, labeling, or scoring. The labels
  distinguish exact system scale from the separate family-specific complexity
  scalar stored with each benchmark item.}
  \label{fig:architecture}
  \par\vspace{1\baselineskip}
  \end{minipage}%
}]

Conditional logistic regressions use 765 observed strata formed by model, task
family, and mechanism. These strata condition out each joint intercept. The
adjusted regression adds log rendered-prompt and canonical expected-answer
character counts; ``unadjusted'' below means unadjusted for these two length
measures, since both regressions are stratified.

Of the 765 observed strata, 423 contain invariant outcomes and therefore
contribute no conditional likelihood. The fits use the remaining 342 strata
and 4{,}841 records while retaining all 18 mechanisms as clusters.
Intervals use a mechanism-cluster CR1 covariance with a \(t_{17}\) reference
distribution. The \(p\)-values for ordinal effects use a two-sided, fully
enumerated Rademacher wild-cluster score test over all \(2^{18}\) sign patterns.

\begin{table}[!t]
\centering
\small
\resizebox{\columnwidth}{!}{%
\begin{tabular}{lrr}
\toprule
Analysis-bin effect & Unadjusted & Adjusted \\
\midrule
B2 vs B1 OR & 0.486 & 0.894 \\
B3 vs B1 OR & 0.259 & 0.734 \\
B4 vs B1 OR & 0.073 & 0.201 \\
\midrule
Ordinal coefficient & -0.822 & -0.479 \\
Ordinal odds ratio & 0.439 & 0.619 \\
95\% OR interval & [0.370, 0.522] & [0.547, 0.701] \\
Wild-score $p$ & $<0.0001$ & $<0.0001$ \\
\bottomrule
\end{tabular}
}
\caption{Exploratory stacked conditional-logistic robustness check for
nine models. Both regressions use strata formed by model, task family, and
mechanism; the adjusted regression adds prompt and expected-answer lengths.
Intervals use CR1--\(t_{17}\) inference, and ordinal \(p\)-values use the fully
enumerated Rademacher wild-cluster score test.}
\label{tab:stacked-accuracy-controls}
\end{table}

Both ordinal fits converged in six and five Newton iterations. For the adjusted
fit, the final score infinity norm is \(4.63\times10^{-13}\), the three-column
information matrix is full rank with condition number 117.0, and the largest
absolute coefficient is 3.73. After conditioning out the invariant strata, the
fitted conditional model triggers no residual separation warning.

This post-hoc result remains exploratory. In particular, the wild score test
has only 18 cluster contributions and relies on their sign-symmetry
approximation; coverage across combinations of family and mechanism is
nonuniform, invariant strata do not identify effects within those strata, and
unrecorded differences in prompt content remain. The shared ordinal coefficient
summarizes the nine evaluated models without establishing equal gradients
across them. The estimate therefore serves as associational sensitivity
evidence and does not support a causal or confirmatory claim.

\FloatBarrier
\subsection{Outcome-Separated Metrics and Baselines}
\label{app:outcome-separated-metrics}

Each model--item outcome is classified as correct; a \emph{semantic failure}
(a payload accepted by normalization that differs from gold); a
\emph{protocol failure} (an empty response or a payload rejected during parsing
or normalization); or an \emph{infrastructure failure} (context rejection,
timeout, gateway, or evaluator failure before comparison). For any reported
slice $D$, let $D_A$ exclude infrastructure failures, $D_J$ contain payloads
accepted by normalization, and $D_C$ contain correct answers. When their
denominators are nonzero, the metrics are
\[
\begin{aligned}
\mathrm{Coverage} &=
  \frac{\lvert D_A\rvert}{\lvert D\rvert},
&\qquad
\mathrm{Compliance} &=
  \frac{\lvert D_J\rvert}{\lvert D_A\rvert},\\[2pt]
\mathrm{Acc}_{\mathrm{semantic}} &=
  \frac{\lvert D_C\rvert}{\lvert D_J\rvert},
&\qquad
\mathrm{Acc}_{\mathrm{e2e}} &=
  \frac{\lvert D_C\rvert}{\lvert D\rvert}.
\end{aligned}
\]
Compliance therefore measures acceptance by the normalizer.

Accuracy remains the primary metric, but it is not the only substantive score.
Fixed label strings use character accuracy; state consistency uses accuracy
over every candidate--scenario consistency decision; episode-graph transitions use
task-level set F1; and trace faithfulness uses count-field accuracy. For family
component score $m_f\in[0,1]$, protocol and infrastructure failures receive zero
in the end-to-end mean
$M_{\mathrm{e2e}}=|D|^{-1}\sum_{r\in D_J}m_f(r)$; the conditional semantic mean
is $M_{\mathrm{semantic}}=|D_J|^{-1}\sum_{r\in D_J}m_f(r)$. Figures use
end-to-end accuracy; the other measures separately identify availability,
protocol compliance, and agreement among valid answers.

Accuracy error bars are 95\% confidence intervals from 10,000 deterministically
seeded, mechanism-level cluster-bootstrap replicates. Entire item clusters are
resampled by mechanism, preserving within-mechanism dependence and item-weighted
accuracy. Shared bootstrap draws across models maintain paired comparisons.
Table~\ref{tab:family-evaluation-controls} reports these metrics and their
bootstrap intervals for GPT-5.5 by task family, alongside the release-derived
majority/constant baseline.

\begin{table*}[t]
\centering
\small
\tabcolsep=2pt
\begin{tabularx}{\textwidth}{>{\raggedright\arraybackslash}X r r r r r r r}
\toprule
Task family & \multicolumn{2}{c}{Majority/constant} & \multicolumn{5}{c}{GPT-5.5} \\
\cmidrule(lr){2-3}\cmidrule(lr){4-8}
 & Accuracy & Partial & E2E accuracy [95\% CI] & E2E partial [95\% CI] & Semantic accuracy & Prot. & Infra. \\
\midrule
Necessity & 0.0 & 50.0 & 23.1 [12.8, 34.8] & 65.6 [59.5, 72.0] & 23.1 & 0 & 0 \\
Plausibility & 0.0 & 50.0 & 36.0 [26.8, 43.8] & 89.2 [86.0, 92.0] & 37.7 & 6 & 0 \\
State consistency & 0.0 & 67.5 & 79.7 [71.9, 87.3] & 98.4 [97.0, 99.3] & 80.1 & 1 & 0 \\
Episode-graph transition & 0.0 & 0.0 & 78.0 [66.2, 94.0] & 88.3 [82.4, 96.6] & 78.0 & 0 & 0 \\
Trace faithfulness & 4.5 & 4.5 & 75.4 [70.1, 81.2] & 75.4 [70.1, 81.2] & 75.4 & 0 & 0 \\
Functional recovery & 11.8 & 37.0 & 62.5 [54.9, 70.1] & 87.3 [84.5, 90.2] & 62.7 & 1 & 0 \\
\bottomrule
\end{tabularx}

\caption{GPT-5.5 outcome-separated metrics and release-derived simple
baselines. Accuracy and partial-score columns are percentages. E2E scores
retain every item; the conditional semantic value reports accuracy among valid
normalized answers. Prot. and Infra. are counts. The family-specific partial
metric is defined in the text.}
\label{tab:family-evaluation-controls}
\end{table*}

\begin{table*}[t]
\centering
\small
\setlength{\tabcolsep}{5pt}
\begin{tabular}{lrrrrrrr}
\toprule
Model & SC & P & N & T & FR & TF & Prot. \\
\midrule
GPT-5.5 & 79.7 & 36.0 & 23.1 & 78.0 & 62.5 & 75.4 & 8 \\
o3 & 25.9 & 5.1 & 6.2 & 57.0 & 46.9 & 43.3 & 52 \\
DeepSeek-R1 & 7.1 & 0.0 & 6.2 & 45.0 & 39.6 & 2.2 & 110 \\
gpt-oss-20b & 0.0 & 0.0 & 1.9 & 39.0 & 29.5 & 0.0 & 152 \\
Qwen3-Coder-30B & 0.9 & 0.0 & 18.1 & 5.0 & 10.4 & 0.0 & 107 \\
GPT-4.1 & 0.5 & 0.0 & 3.1 & 15.0 & 12.8 & 0.0 & 24 \\
Llama-3.3-70B & 0.5 & 0.0 & 15.6 & 1.0 & 9.7 & 0.0 & 255 \\
Qwen3-30B & 3.3 & 0.0 & 0.6 & 15.0 & 7.6 & 0.0 & 441 \\
Gemma-3-12B & 0.0 & 0.0 & 5.0 & 0.0 & 6.2 & 0.0 & 147 \\
\bottomrule
\end{tabular}
\caption{End-to-end accuracy (\%) by model and task family. SC denotes state consistency, P plausibility, N necessity, T episode-graph transitions, FR functional recovery, and TF trace faithfulness. Prot. is the total number of protocol failures across all 1{,}120 items.}
\label{tab:family-model-metrics}
\end{table*}

The majority/constant baseline repeats the release-wide majority label for
fixed strings, predicts every candidate inconsistent for state consistency,
uses the empty transition set, or predicts the modal trace count. It is fitted
to the released gold-label marginal and is therefore a diagnostic floor rather
than a held-out learning baseline. Deterministic replay of the structured
generator state obtains 100\% accuracy by construction and serves only as an
integrity check. The prompt-text symbolic parser is the non-oracle, prompt-only
systems baseline. It reconstructs the typed task from the rendered prompt and
executes the same public qualitative solver and semantics used elsewhere in the
release. Its inference path reads no solutions, catalog files, generator
objects, or simulator caches. It parses and solves all 1{,}120 items with 100\%
accuracy.

\paragraph{Complete per-model, per-family report.}
Table~\ref{tab:family-model-metrics} reports end-to-end accuracy for every model
and family together with each model's total protocol failures.
Ten Qwen3-30B requests (two plausibility and eight functional
recovery) returned empty responses and are classified as protocol
failures. Protocol failures total 1{,}296.

\subsection{Operational Semantic Definitions}
\label{app:formal-properties}

The operational semantic clauses expand the compact definitions from the main
paper. They fix the solved objects and exact
admissibility and labeling conditions used by the benchmark.

\begin{definition}[Compiled state and admissible interpretation]
For $\M$, $T_{p,\sigma}$ is the finite set of declared outgoing transition
records for component $p$ in local state $\sigma$. A record $t$ gives a
destination state, an optional event coordinate, and finite source and target
guard sets $H_t^-$ and $H_t^+$ of pairs $(v,A)$ with $A\subseteq Q(v)$. We write
$\iota\models H$ when $\iota(v)\in A$ for every $(v,A)\in H$. These transition
records do not constrain $\Iset_{s,r}$; the separate simulation session
evaluates them only when constructing episode-graph edges.

Given a composite component-state assignment $s \in S$ and a scenario
$r \in B$, the compiler restricts each variable to a domain
$D_{s,r}(v)\subseteq Q(v)$ using the active state constraints and scenario
restrictions. It also instantiates the active qualitative constraints
\[
C(s,r)=\operatorname{Inst}_{s,r}(E,N)
\]
where $\operatorname{Inst}_{s,r}$ selects the active component-local
confluences and connection laws, renames them to their component instances, and
applies their state and scenario conditions. Each $c\in C(s,r)$ is a finite
list of signed product terms $u_1,\ldots,u_m$. Let
$\operatorname{qsgn}_{\iota}(u_j)\in\{-,0,+\}$ be the qualitative sign obtained
by multiplying the coefficient sign and the signs under $\iota$ of all factors,
with their exponents respected. Confluence satisfaction is
\[
\iota\models c
\iff
\left[\forall j,\ \operatorname{qsgn}_{\iota}(u_j)=0\right]
\ \lor\
\left[
\begin{aligned}
&\exists j,\ \operatorname{qsgn}_{\iota}(u_j)=+,\\[-1mm]
&\exists k,\ \operatorname{qsgn}_{\iota}(u_k)=-
\end{aligned}
\right],
\]
and $\iota\models C(s,r)$ iff $\iota\models c$ for every $c\in C(s,r)$.
Thus a zero term together with nonzero terms of only one sign does not satisfy a
confluence. An \emph{admissible interpretation} is an assignment $\iota$ of
qualitative values to every active variable in $V$ such that
$\iota(v)\in D_{s,r}(v)$ for every $v$ and all constraints in $C(s,r)$ are
satisfied. The intrastate solver returns the finite set
\[
\begin{aligned}
\Iset_{s,r}=\{\iota :{}& \iota(v)\in D_{s,r}(v)\text{ for every }v,\\
& \iota\models C(s,r)\}.
\end{aligned}
\]
\end{definition}

\paragraph{Operational edge-admission clauses and bounded reachability.}
The episode-graph definition in the main paper fixes the episode universe
$\mathcal V_r$ and graph $\mathcal G_r=(\mathcal V_r,\mathcal A_r)$. The
following clauses specify the deterministic edge-admission procedure used by
the released simulator.

For a source episode $x=(s,\iota)$, the constructor derives a cause set
$\mathcal C_r(x)$ from the active laws, derivative coordinates, and quantity
spaces. A boundary cause is an atomic set of nearest directed termination events
after exact coincidence, equality-change, and epsilon ordering. An intrastate
cause has an empty event set and is available only when no mandatory point-cell
departure preempts it.

For episodes $x,x'\in\mathcal V_r$ and $c\in\mathcal C_r(x)$, the exact endpoint
solver returns the Boolean $\operatorname{EdgeOK}_r(x,x',c)$. It returns true
exactly when all of the following implementation clauses pass:
\begin{enumerate}
\item[\textsc{E1}] Each derivative coordinate remains fixed unless a selected
event moves it to that event's target cell; parameters and nondynamic inputs
remain fixed.
\item[\textsc{E2}] Every boundary event reaches the nearest cell in its
directed quantity space.
\item[\textsc{E3}] Coincident groups, mandatory equality changes, and epsilon
ordering are respected.
\item[\textsc{E4}] Every changed non-derivative quantity has either an
initiating event or a causal predecessor under an active law.
\item[\textsc{E5}] The target $x'$ is a solved interpretation of its composite
state.
\item[\textsc{E6}] Every continuous quantity remains in the same qualitative
cell or moves to an adjacent cell.
\item[\textsc{E7}] Every coordinate change agrees with the source value of its
derivative under the qualitative mean-value rule.
\item[\textsc{E8}] Feedback may propagate an event-seeded change but may not
initiate a change without an event-seeded causal path.
\end{enumerate}

Declared transitions constrain every component-state change. For
$x=(s,\iota)$ and $x'=(s',\iota')$, a declaration $t$ is compatible with a
change of component $p$ when
\[
t\in T_{p,s(p)},\quad
\operatorname{dst}(t)=s'(p),\quad
\iota\models H_t^-,\quad
\iota'\models H_t^+,
\]
and its event coordinate, when present, occurs in $c$. Let
$\operatorname{Decl}(x,x',c)$ mean that every changed component has such a
compatible declaration. An unchanged component needs no declaration. The
admissible episode relation is
\[
\mathcal A_r=
\left\{
(x,x',c)\ \middle|\
\begin{aligned}
&x,x'\in\mathcal V_r,\quad x\ne x',\\
&c\in\mathcal C_r(x),\\
&\operatorname{EdgeOK}_r(x,x',c),\\
&\operatorname{Decl}(x,x',c)
\end{aligned}
\right\}.
\]
For an initial episode $x_0$, define the exact-depth episode and edge layers by
\[
\begin{aligned}
R_0(x_0)&=\{x_0\},\\
\mathcal A_r^{(d)}(x_0)
  &=\{(x,x',c)\in\mathcal A_r:x\in R_{d-1}(x_0)\},\\
R_d(x_0)
  &=\{x':\exists x,c,\ (x,x',c)\in\mathcal A_r^{(d)}(x_0)\},
  \qquad d\geq 1.
\end{aligned}
\]
The bounded reachable edge set through horizon $H$ is
$\mathcal A_r^{\leq H}(x_0)=
\bigcup_{d=1}^{H}\mathcal A_r^{(d)}(x_0)$.

\begin{definition}[State-consistency label]
Let $r_k$ be the scenario formed by combining the persistent operating-context
restrictions with any additional restrictions displayed for scenario $k$, and
let $s_a$ be candidate row $a$'s complete composite state. Then
\[
L_{\mathrm{cons}}(k,a)=\Yes
\iff \Iset_{s_a,r_k}\ne\varnothing.
\]
\end{definition}

\begin{definition}[Bounded-path plausibility]
Let $x_0$ be the supplied initial episode, let $H$ be the successor horizon,
and let $O_{q,t}$ be the conjunction of observations in narrative $q$ at
checkpoint $t$. Then
\[
L_{\mathrm{plaus}}(q)=\mathrm{P}
\iff \exists x_1,\ldots,x_H,\ c_1,\ldots,c_H
\]
such that, for every $t=1,\ldots,H$,
$(x_{t-1},x_t,c_t)\in\mathcal A_r$ and $x_t\models O_{q,t}$.
Otherwise $L_{\mathrm{plaus}}(q)=\mathrm{I}$. In particular, all checkpoint
observations in one narrative must be witnessed by the same exact-length path.
\end{definition}

\begin{definition}[Edge-graph necessity]
Let $\widehat{\mathcal A}$ be the exact reachable edge set supplied in the task.
A claim
$q$ contains a conjunction of source facts $\alpha_q$ and target alternatives
$\beta_{q,1},\ldots,\beta_{q,m}$. Define its antecedent-support and
counterexample sets as
\[
\operatorname{Supp}(q)=
\{(x,x',c)\in\widehat{\mathcal A}:x\models\alpha_q\}
\]
and
\[
\begin{aligned}
\operatorname{CE}(q)=\{(x,x',c)\in\widehat{\mathcal A}:{}&
x\models\alpha_q,\\
&x'\not\models\beta_{q,j}\text{ for every }j\}.
\end{aligned}
\]
A supported benchmark claim must satisfy
$\operatorname{Supp}(q)\ne\varnothing$. On this eligible domain, the label is
$\mathrm{N}$ exactly when $\operatorname{CE}(q)=\varnothing$ and $\mathrm{U}$
otherwise. Necessity is therefore evaluated over the supplied reachable edges,
not over independent interpretations.
\end{definition}

\begin{definition}[Witnessed component-state transitions]
Let $x_0$ and $H$ be the supplied initial episode and horizon. For
$x=(s,\iota)$ and $x'=(s',\iota')$, the transition task returns the sorted set
\[
\begin{aligned}
\mathcal T_H(x_0)=\{(p,s(p),s'(p)):{}&
(x,x',c)\in\mathcal A_r^{\leq H}(x_0),\\
&p\in P,\quad s(p)\ne s'(p)\}.
\end{aligned}
\]
Thus every returned triple is witnessed by an admissible reachable edge.
Unrealized declarations and interpretation-only edges are excluded.
\end{definition}

\begin{definition}[Functional-recovery status]
For a repair policy $\pi$, let $O_\pi$ be its complete named outcome set and
let
\[
\operatorname{Succ}(o)=
\{x\in\mathcal V_r:\exists c\in\mathcal C_r(o),\
(o,x,c)\in\mathcal A_r\}
\]
be the targets of all admissible one-edge episode transitions from outcome
$o$. These are the prompt's ``physical episode edges'': they include
interpretation-only edges between distinct episode nodes with the same composite
state, but exclude unchanged stuttering. The one-edge (episode-depth-one)
deadline set is
\[
\operatorname{Dead}(o)=
\begin{cases}
\operatorname{Succ}(o),&\operatorname{Succ}(o)\ne\varnothing,\\
\{o\},&\operatorname{Succ}(o)=\varnothing.
\end{cases}
\]
The second branch is a contract-only hold for the deadline check, not a physical
episode edge.
Writing $\mathsf{Unsafe}$ and $\mathsf{Goal}$ for the supplied conjunctions,
call $\pi$ unsafe if any $o\in O_\pi$ or any
$x\in\operatorname{Succ}(o)$ satisfies $\mathsf{Unsafe}$. If it is not unsafe,
call it deadline-failing if some $x\in\operatorname{Dead}(o)$ for some
$o\in O_\pi$ does not satisfy $\mathsf{Goal}$. The policy status is
\[
L_{\mathrm{rec}}(\pi)=
\begin{cases}
\mathrm{U},&\pi\text{ is unsafe},\\
\mathrm{D},&\pi\text{ is deadline-failing},\\
\mathrm{R},&\text{otherwise}.
\end{cases}
\]
The task returns one such $\mathrm{U}/\mathrm{D}/\mathrm{R}$ status in displayed
policy order.
\end{definition}

\begin{definition}[Trace-faithfulness count]
Let the independent displayed traces be indexed by
$g=1,\ldots,n_{\mathrm{tr}}$. Trace $g$
starts from the initial working domains $D_{g,0}$ supplied by its referenced
context and contains $h_g$ displayed steps. For step $j$ of trace $g$, let
$b_{g,j}$ and $a_{g,j}$ be its displayed before- and after-sets, $v_{g,j}$ its
target, and $c_{g,j}$ its cited confluence. If
$\operatorname{Support}(c_{g,j},D_{g,j-1},v_{g,j})$ is the exact set of target
cells supported by assignments to the other variables in $c_{g,j}$, write
$S_{g,j}$ for this set. Then
\[
\begin{aligned}
L_{\mathrm{trace}}(g,j)=\Yes\iff{}&
b_{g,j}=D_{g,j-1}(v_{g,j})\\
&{}\wedge v_{g,j}\in c_{g,j}\\
&{}\wedge S_{g,j}=a_{g,j}.
\end{aligned}
\]
Steps within each trace are processed in order, with
\[
D_{g,j}=D_{g,j-1}[v_{g,j}\mapsto a_{g,j}]
\]
even when the citation is unfaithful. Each trace resets independently to its
own $D_{g,0}$. The task returns
\[
\sum_{g=1}^{n_{\mathrm{tr}}}\sum_{j=1}^{h_g}
\mathbf{1}\!\left[L_{\mathrm{trace}}(g,j)=\No\right],
\]
the total number of unfaithful displayed steps across all traces.
\end{definition}

\subsection{Candidate Construction, Validation, and Rendering}
\label{app:candidate-construction}

All six generators consume validated catalog mechanisms and exact simulator
objects; no evaluated LLM participates in candidate construction, labeling, or
selection. Each generator first creates a deterministic semantic candidate
pool, computes its gold object with the corresponding exact solver, and only
then renders the user-facing task. Table~\ref{tab:candidate-construction}
records the family-specific construction and admission checks frozen for
v1.0.0.

\begin{table*}[!t]
\centering
\small
\tabcolsep=3pt
\begin{tabularx}{\textwidth}{
  >{\raggedright\arraybackslash}p{2.25cm}
  >{\raggedright\arraybackslash}X
  >{\raggedright\arraybackslash}X}
\toprule
Family & Candidate construction & Semantic admission and fixed controls \\
\midrule
State consistency &
Complete composite-state rows are drawn from the mechanism state space and
paired with persistent context plus scenario-specific restrictions. Inconsistent
rows are deliberately not prefiltered. &
The exact intrastate relation is queried for every candidate--scenario pair.
Reachability and transition rules are outside this family. Retained candidates
must fall in the frozen absolute D1--D4 constraint-count bands. \\
Plausibility &
One complete initial episode and its exact bounded graph yield 32
checkpoint narratives. Positive packets come from one complete path and
layer invariants; negative packets are constructed against exact
non-witness certificates. &
Every narrative contains three to six facts at every checkpoint, with 16
\(\mathrm P\) and 16 \(\mathrm I\) labels. All facts in a positive narrative
share one exact-length witness. Horizons are 1, 1, 2, and 3 for D1--D4. \\
Necessity &
Four claims are mined over one exact reachable edge set. Antecedents contain
three or four source facts and consequents contain two or three distinct target
alternatives. &
Antecedents are nonvacuous. Necessary claims require the joint disjunction and
reject a sufficient proper disjunction; unnecessary claims retain an exact
counterexample, including one at the deepest expanded source layer. The normal
task has a two--two \(\mathrm N/\mathrm U\) balance. \\
Episode-graph transitions &
A complete initial episode and horizon determine an exact bounded graph; each
candidate answer projects distinct component-state changes witnessed by its
edges. Seeded surface variants change evidence order only. &
The answer must be nonempty. Unrealized declarations and interpretation-only
edges are excluded, while simultaneous component changes are projected
individually. Preferred horizons are 1, 1, 2, and 2 for D1--D4. \\
Functional recovery &
Removing the declared initial-coordinate restrictions produces the exhaustive
fault-belief episode library for the task. Externally observable goal and
unsafe atoms define contracts, and each policy names its complete repaired
outcome set. &
Labels quantify over every named outcome and every exact one-edge successor.
D1--D4 display 1, 3, 5, and 15 plans; D4 contains five each of
\(\mathrm U,\mathrm D,\mathrm R\), and policy order is seeded before taking the
level prefix. \\
Trace faithfulness &
Exact local confluence support generates six-step sequential traces. Faithful
steps use the exact before-domain and supported after-set; unfaithful steps
apply a controlled wrong law, before-domain mismatch, or near-miss supported
set and are rechecked by the same exact solver. &
Trace scenarios are independent but steps within a trace update sequentially.
D1--D4 contain 22, 29, 37, and 44 traces, hence 132, 174, 222, and 264
displayed step judgments. \\
\bottomrule
\end{tabularx}
\caption{Frozen task-family candidate construction and semantic admission
rules.}
\label{tab:candidate-construction}
\end{table*}

\paragraph{Shared validation and selection.}
For every candidate, the generator checks the family answer shape and
invariants, recomputes the answer-independent complexity scalar, assigns the
frozen D1--D4 level. The release targets four tasks per supported
family--mechanism--level cell. If exact search or the prompt ceiling prevents
filling a cell, the cell remains partial or unsupported rather than receiving
a substitute task from another level. Selection uses deterministic
ordering and seeded tie breaks; after cell aggregation, normalized duplicate
prompts are removed.

\paragraph{Rendering and persistence.}
The family renderer serializes only the mechanism evidence, semantic candidate,
instructions, and requested answer shape. Gold objects and audits are persisted
separately from task content. Stable task identifiers and completion hashes bind
the catalog, generator settings, task, and solution artifacts, so a
change in semantic content or rendering invalidates the recorded completion
state.

\subsection{Scenario-Level Simulator Protocol}
\label{app:scenario-protocol}

Algorithm~\ref{alg:scenario-protocol} expands the construction of the shared
semantic objects \(Z_M\) invoked by Algorithm~1 in the main paper and makes the
two stages of the released v1.0.0 implementation explicit. The intrastate
solver returns an active model and a factorized satisfying relation; its
\texttt{ActiveModel} contains selected states, domains, active confluences, and
qualitative signs, but no transitions. The simulation session separately builds
the episode universe, collects $T(s)$ for each source state, and applies
declaration, guard, cause, continuity, and endpoint checks through the exact
endpoint solver to construct the bounded reachable graph. Small relations may be
materialized as a cache, but exact counts and queries operate on the factors or
their decision diagrams. Each constrained component receives exact
generalized-arc propagation on bounded-scope laws; decision-diagram compilation
then prunes impossible prefixes and merges equal confluence residuals.

\begin{algorithm}
\caption{Factorized state solving and separate episode-graph construction}
\label{alg:scenario-protocol}
\begin{algorithmic}[1]
\Require mechanism $\M$, complete state $s$, restrictions $r$; optionally
initial episodes $X_0$ and horizon $H$
\Statex \textbf{Intrastate solve}
\State $\bar s\gets\operatorname{NormalizeComplete}(s)$
\State $D\gets\{v\mapsto\operatorname{Values}(Q_v):v\in V\}$; fix parameter cells
\State $C\gets\operatorname{ActiveLaws}(\M,\bar s)$
\State $D\gets\operatorname{Restrict}(D,\operatorname{ActiveStateRestrictions}(\M,\bar s)\cup r)$
\State $\mathcal A\gets(\bar s,D,C,\operatorname{QualitativeSigns}(\M))$
\State $\mathcal F\gets[\,]$
\ForAll{$K\in\operatorname{ConnectedComponents}(D,C)$, cheapest first}
  \State $D_K^\star\gets\operatorname{GAC}_{\mathrm{bounded}}(D_K,C_K)$
  \If{some domain in $D_K^\star$ is empty}
    \State \textbf{return} $\mathcal A$, the empty relation, and count $0$
  \EndIf
  \State $\pi_K\gets\operatorname{ReverseMinFill}(D_K^\star,C_K)$
  \State $B_K\gets\operatorname{CompileReducedMDD}(\pi_K,D_K^\star,C_K)$
  \If{$\operatorname{Count}(B_K)=0$}
    \State \textbf{return} $\mathcal A$, the empty relation, and count $0$
  \EndIf
  \State append $B_K$ to $\mathcal F$ (store tuples instead when small)
\EndFor
\State $\widehat{\Iset}_{s,r}\gets\mathop{\times}_{K}\mathcal F_K$;
$|\Iset_{s,r}|\gets\prod_K\operatorname{Count}(\mathcal F_K)$
\State \textbf{return from solver:} $\mathcal A$, $\mathcal F$,
$|\Iset_{s,r}|$, and necessary facts
\Statex \textbf{Episode construction (separate simulation session)}
\ForAll{complete composite states $s'$}
  \State solve $(s',r)$ and retain each nonempty factor relation
\EndFor
\State $\mathcal V_r\gets
\bigcup_{s'}\{(s',\iota):\iota\in\Iset_{s',r}\}$
\State $\mathcal T_r\gets
\{s'\mapsto T(s'):s'\text{ occurs in }\mathcal V_r\}$
\State $\mathcal O_r\gets
\operatorname{ExactEndpointSolver}(\mathcal V_r,\mathcal T_r)$ with
cause, causal, continuity, declaration, guard, and endpoint checks
\State $\mathcal A_r^{\leq H}(X_0)\gets
\operatorname{PairedDDExpand}(X_0,H,\mathcal O_r)$
\State \textbf{return from session:} $\mathcal V_r$, $\mathcal O_r$, and
$\mathcal A_r^{\leq H}(X_0)$
\end{algorithmic}
\end{algorithm}

\subsection{Prompt and Evaluator Contract}
\label{app:prompt-evaluator}

The item prompt is generated once per task by the task-family renderer and is
stored as the task content. Every evaluation request uses the same two-message
envelope: the fixed system message reproduced in each excerpt below, followed
by the persisted task content as the user message. No task-family schema is sent
out of band. Endpoints whose profiles support a structured response constraint
receive only generic JSON-object mode. Each backend receives one request;
model-directed tools, repository access, file inspection, and command execution
are outside the comparison protocol.

\subsubsection{Request Controls}

The persisted direct-inference model runs use a shared evaluation protocol
that sets: concurrency 20, a 600-second request timeout, a required
input capacity of at least 80{,}000 tokens, a target output limit of 20{,}000
tokens, and low reasoning effort only for profiles that expose that
control. The effective output limit is the smaller of 20{,}000 and the
profile's declared maximum. Thus the limit is a request ceiling, not a claim
about the number of tokens actually produced. Temperature zero is sent only to
profiles that support temperature; it is omitted otherwise.

\subsubsection{Current Rendered Prompt Excerpts}

The following boxes reproduce paper-sized excerpts from persisted v1.0.0
prompts. Omissions from long evidence tables are marked explicitly; the task
files retain the complete rendered text used during evaluation.

\begin{promptbox}[State Consistency]
\promptsection{System}
Answer qualitative mechanistic benchmark tasks as JSON. Return only the
requested JSON object.

\promptsection{User}
\begin{promptmeta}
\promptfield{Task family}{\mechex{state consistency}}
\promptfield{Mechanism id}{\mechex{mechanics\_first}}
\promptfield{Task id}{\mechex{task\_f1024887d48f01c9}}
\promptfield{Prompt tokens}{\mechex{9{,}380}}
\end{promptmeta}

\promptsection{Rendered Prompt Excerpt}
\textcolor{taskprompttemplate}{Audit complete composite-state consistency for
this qualitative mechanism.}

\begin{promptitems}
\item \textbf{Semantics:} Each candidate row selects one state for every
multi-state component and uses the sole state of every other component. A row
is consistent in a case exactly when at least one complete joint assignment of
qualitative cells satisfies all selected-state restrictions, the persistent
context, that case's restrictions, and every active local and connection law.
Persistent background restrictions explicitly include the exact qualitative
cell of every displayed bound parameter.
Transition reachability is out of scope.
\item \textbf{Component-state clauses:}
\mechex{\code{damper} has \code{ROTATING_NEGATIVE}, \code{REST}, and
\code{ROTATING_POSITIVE}; \code{inertia1}, \code{inertia2}, and
\code{inertia3} have the same three local states; \code{spring} has
\code{NEGATIVE_LOAD}, \code{UNLOADED}, and \code{POSITIVE_LOAD}.}
\item \textbf{Cases, evaluated independently:}
\mechex{\code{C1}: \code{inertia3.w} in \{positive\}; \code{C2}:
\code{inertia1.w} in \{negative\}; \code{base}: no additional restrictions;
\code{C3}: \code{inertia2.w} in \{negative\}.}
\item \textbf{Candidate complete composite states:}
\mechex{\code{S001}: \code{REST | REST | REST | REST | UNLOADED};
\code{S002}: \code{ROTATING_NEGATIVE | ROTATING_POSITIVE |
ROTATING_NEGATIVE | ROTATING_NEGATIVE | POSITIVE_LOAD}; \code{S003}:
\code{ROTATING_NEGATIVE | ROTATING_POSITIVE | ROTATING_POSITIVE |
ROTATING_NEGATIVE | POSITIVE_LOAD};
\placeholder{S004 through S016}.}
\item \textbf{Additional task data:}
\mechex{\placeholder{remaining clauses, variables, restrictions, and
confluences}.}
\end{promptitems}

\textcolor{taskprompttemplate}{Qualitative confluence rule: determine each
signed term after multiplying its coefficient and factor signs. A zero-sum
confluence is satisfiable exactly when all terms are zero, or when at least one
positive and one negative term occur. A zero term alongside only positive
terms, or alongside only negative terms, does not balance the confluence.}

\textcolor{taskprompttemplate}{For every case, return the sorted consistent
candidate ids and their count.}

\promptsection{Required Output Format}
\begin{prompttemplateblock}
\{"answer": \{"case\_results": \{\mechex{"C1"}:
\{"consistent\_count": 0, "consistent\_ids": []\},
\mechex{"C2"}: \{"consistent\_count": 0, "consistent\_ids": []\},
\mechex{"C3"}: \{"consistent\_count": 0, "consistent\_ids": []\},
\mechex{"base"}: \{"consistent\_count": 0, "consistent\_ids": []\}\}\},
"rationale": ""\}
\end{prompttemplateblock}
\end{promptbox}

\begin{promptbox}[Plausibility]
\promptsection{System}
Answer qualitative mechanistic benchmark tasks as JSON. Return only the
requested JSON object.

\promptsection{User}
\begin{promptmeta}
\promptfield{Task family}{\mechex{plausibility}}
\promptfield{Mechanism id}{\mechex{mechanics\_accelerate}}
\promptfield{Task id}{\mechex{task\_376eb16721d408da}}
\promptfield{Prompt tokens}{\mechex{11{,}948}}
\end{promptmeta}

\promptsection{Rendered Prompt Excerpt}
\textcolor{taskprompttemplate}{Classify bounded episode-history plausibility
for a qualitative mechanism.}

\textcolor{taskprompttemplate}{The scenario supplies one complete initial
episode and a successor horizon H. Derive the exact reachable episode graph
through H transitions. Each candidate narrative gives conjunctive observations
at every checkpoint t=1 through H. It is \code{plausible} exactly when one path
of exactly H admissible edges starts at that episode and satisfies every
observation at checkpoint t in the episode reached after exactly t edges. All
observations in one narrative must use the same path.}

\begin{promptitems}
\item \textbf{Episode-edge rules:} Nonzero derivatives at points make
departure immediate and mandatory; interval departures are optional. Mandatory
equality changes suppress finite optional exits. Targets must satisfy declared
component-state transitions and guards, event-subset, causality, continuity,
and qualitative mean-value constraints. Do not add unchanged stuttering steps.
\item \textbf{Model and scenario data:}
\mechex{The initial composite state is
\code{accelerate.unconditional}, \code{constantAcc.unconditional}, and
\code{mass.REST}; the successor horizon is \code{1};
\code{accelerate.a=positive}, \code{accelerate.v=zero},
\code{constantAcc.y=positive}, and \code{mass.s=positive};
\placeholder{remaining compact qualitative-model, context, and initial-episode
JSON}.}
\item \textbf{Candidate narratives:}
\mechex{\code{Q01}: t=1:
\code{accelerate.d1__flange__f = positive} AND
\code{mass.d2__flange_b__s = positive} AND
\code{mass.flange_a.s = zero}; \code{Q02}: t=1:
\code{accelerate.d2__flange__s = positive} AND
\code{accelerate.s = zero} AND \code{mass.flange_a.s = zero};
\code{Q03}: t=1: \code{accelerate.flange.s = zero} AND
\code{mass.a = positive} AND \code{mass.s = positive};
\placeholder{Q04 through Q32}.}
\end{promptitems}

\textcolor{taskprompttemplate}{Write labels in query-id order Q01 through Q32.
Use \code{P} for plausible and \code{I} for implausible. The label string must
contain exactly 32 uppercase P/I characters, with no spaces, separators, or
other letters. Replace all 32 placeholders and return only this compact JSON.}

\promptsection{Required Output Format}
\begin{prompttemplateblock}
\{"answer":\{"plausibility\_labels":
"XXXXXXXXXXXXXXXX\allowbreak{}XXXXXXXXXXXXXXXX"\}\}
\end{prompttemplateblock}
\end{promptbox}

\begin{promptbox}[Necessity]
\promptsection{System}
Answer qualitative mechanistic benchmark tasks as JSON. Return only the
requested JSON object.

\promptsection{User}
\begin{promptmeta}
\promptfield{Task family}{\mechex{necessity}}
\promptfield{Mechanism id}{\mechex{electrical\_resistor}}
\promptfield{Task id}{\mechex{task\_4e4e78586f6ee210}}
\promptfield{Prompt tokens}{\mechex{17{,}741}}
\end{promptmeta}

\promptsection{Rendered Prompt Excerpt}
\textcolor{taskprompttemplate}{Classify bounded temporal necessity claims.}

\textcolor{taskprompttemplate}{The scenario supplies the complete exact
in-scope episode-edge graph produced by the qualitative simulator. It is
losslessly normalized so repeated source interpretations, component states,
and evidence are stored once. All indices are zero-based; node and edge
ordinals are implicit from their supplied row order.}

\begin{promptitems}
\item \textbf{Scenario data:}
\mechex{The successor horizon is \code{1}. The normalized graph contains
\code{65} nodes, \code{64} distinct edges, and \code{1} cause over
\code{58} aliased variables, \code{5} components, and \code{6} states;
\placeholder{compact aliased model, operating context, and normalized graph
rows}.}
\item \textbf{Claim semantics:} Each retained claim has at least one admissible
edge whose source satisfies all supplied SOURCE facts. Each claim has SOURCE
facts and a disjunction of TARGET facts. It is \code{necessary} exactly when
every such edge has a target satisfying at least one TARGET alternative. A zero
exact counterexample count means \code{necessary}; any positive count means
\code{not_necessary}.
\item \textbf{Claims:}
\mechex{\code{Q01}: IF SOURCE \code{v1 = zero} AND SOURCE
\code{v36 = zero} AND SOURCE \code{v44 = zero} AND SOURCE
\code{v50 = positive}, THEN TARGET \code{v8 = zero} OR TARGET
\code{v37 = negative}; \code{Q02}: IF SOURCE \code{v3 = zero} AND SOURCE
\code{v14 = zero} AND SOURCE \code{v48 = positive}, THEN TARGET
\code{v35 = zero} OR TARGET \code{v37 = negative}; \code{Q03}: IF SOURCE
\code{v2 = zero} AND SOURCE \code{v12 = zero} AND SOURCE \code{v42 = zero}
AND SOURCE \code{v54 = positive}, THEN TARGET \code{v0 = zero} OR TARGET
\code{v5 = negative}; \code{Q04}: IF SOURCE \code{v1 = zero} AND SOURCE
\code{v36 = zero} AND SOURCE \code{v44 = zero} AND SOURCE
\code{v50 = positive}, THEN TARGET \code{v8 = positive} OR TARGET
\code{v37 = negative}.}
\end{promptitems}

\textcolor{taskprompttemplate}{Audit every claim against the supplied edge
rows. Return \code{N} when the exact counterexample count is zero and \code{U}
when it is positive. Write labels in query-id order Q01 through Q04. Replace
each \code{X} with \code{N} or \code{U} and return only this compact JSON
object.}

\promptsection{Required Output Format}
\begin{prompttemplateblock}
\{"answer":\{"necessity\_labels":"XXXX"\}\}
\end{prompttemplateblock}
\end{promptbox}

\begin{promptbox}[Episode-Graph Transitions]
\promptsection{System}
Answer qualitative mechanistic benchmark tasks as JSON. Return only the
requested JSON object.

\promptsection{User}
\begin{promptmeta}
\promptfield{Task family}{\mechex{transition}}
\promptfield{Mechanism id}{\mechex{mechanics\_accelerate}}
\promptfield{Task id}{\mechex{task\_ae9a1a1f30629451}}
\promptfield{Prompt tokens}{\mechex{10{,}564}}
\end{promptmeta}

\promptsection{Rendered Prompt Excerpt}
\textcolor{taskprompttemplate}{List every admissible component-state transition
in this mechanism scenario.}

\textcolor{taskprompttemplate}{The scenario supplies one complete initial
episode and a successor horizon H. Derive the exact reachable episode graph
through depth H and return the distinct component-state changes realized by at
least one admissible edge whose source depth is less than H. Include edges
entering depth H, but no outgoing edges from depth H. Return semantic triples,
not episode-edge identifiers.}

\begin{promptitems}
\item \textbf{Transition rules:} Mandatory point departures suppress finite
optional exits; otherwise only compatible optional event subsets are allowed.
Every changed component must match a declared source/target state pair, its
guards, and its event coordinate when declared. Targets must satisfy
consistency, causality, continuity, and the qualitative mean-value rule.
Interpretation-only edges must not be returned.
\item \textbf{Model and scenario data:}
\mechex{The initial composite state is
\code{accelerate.unconditional}, \code{constantAcc.unconditional}, and
\code{mass.REST}; the successor horizon is \code{1};
\code{accelerate.a=positive}, \code{accelerate.v=zero},
\code{constantAcc.y=positive}, and \code{mass.s=positive};
\placeholder{remaining compact qualitative-model, context, and initial-episode
JSON}.}
\end{promptitems}

\textcolor{taskprompttemplate}{If one edge changes several components, return
one triple per component. Deduplicate triples witnessed by multiple edges or
causes. Return the exact list sorted by \code{component}, then
\code{source_state}, then \code{target_state}.}

\promptsection{Required Output Format}
\begin{prompttemplateblock}
\{"answer": \{"admissible\_component\_state\_transitions": []\},
"rationale": ""\}
\end{prompttemplateblock}
\end{promptbox}

\begin{promptbox}[Functional Recovery]
\promptsection{System}
Answer qualitative mechanistic benchmark tasks as JSON. Return only the
requested JSON object.

\promptsection{User}
\begin{promptmeta}
\promptfield{Task family}{\mechex{functional recovery}}
\promptfield{Mechanism id}{\mechex{mechanics\_accelerate}}
\promptfield{Task id}{\mechex{task\_53fd103754d4f659}}
\promptfield{Prompt tokens}{\mechex{10{,}737}}
\end{promptmeta}

\promptsection{Rendered Prompt Excerpt}
\textcolor{taskprompttemplate}{Classify one-step temporal functional recovery
for a qualitative mechanism.}

\textcolor{taskprompttemplate}{The displayed fault-belief episode library is
the complete uncertainty set for this task; do not add other catalog episodes.
Each repair policy names its complete possible post-repair episode ids.
Evaluate policies independently and universally over every named outcome. For
each outcome, derive every admissible successor after exactly one physical
episode edge. If the outcome has no successor, hold that outcome for the
deadline check only. This contract hold is not a physical episode edge.}

\begin{promptitems}
\item \textbf{Fault definition:}
\mechex{Basis \code{loss_of_declared_initial_coordinate_restrictions}; lost
restrictions \code{accelerate.s} in \{zero\} and
\code{accelerate.v} in \{zero\}.}
\item \textbf{Fault-belief episode library:}
\mechex{\code{E01} has composite state
\code{accelerate.unconditional}, \code{constantAcc.unconditional}, and
\code{mass.MOVING_NEGATIVE}; selected cells include
\code{accelerate.d1__flange__f=negative}, \code{accelerate.s=zero},
\code{accelerate.v=negative}, \code{mass.s=positive}, and
\code{mass.v=negative}; \placeholder{remaining episode cells}.}
\item \textbf{Recovery contract:}
\mechex{Horizon and deadline are both \code{1};
\code{goal_atoms} requires
\code{accelerate.d1__flange__f=negative}; \code{unsafe_atoms} is
\code{accelerate.d1__flange__f=zero}.}
\item \textbf{Repair policies:}
\mechex{\code{R01} names the outcome episode \code{E01}.}
\item \textbf{Additional task data:}
\mechex{\placeholder{model_and_context_JSON}.}
\end{promptitems}

\textcolor{taskprompttemplate}{Status priority for each policy: \code{U} if
any outcome or successor satisfies \code{unsafe_atoms}; otherwise \code{D} if
any deadline episode fails \code{goal_atoms}; otherwise \code{R}. The function
is only the external goal/unsafe contract.}

\textcolor{taskprompttemplate}{Write one status character in displayed policy
order. Replace all \mechex{1} placeholders and return only this compact JSON.}

\promptsection{Required Output Format}
\begin{prompttemplateblock}
\{"answer":\{"plan\_statuses":\mechex{"X"}\}\}
\end{prompttemplateblock}
\end{promptbox}

\begin{promptbox}[Trace Faithfulness]
\promptsection{System}
Answer qualitative mechanistic benchmark tasks as JSON. Return only the
requested JSON object.

\promptsection{User}
\begin{promptmeta}
\promptfield{Task family}{\mechex{trace faithfulness}}
\promptfield{Mechanism id}{\mechex{thermal\_two\_masses}}
\promptfield{Task id}{\mechex{task\_df5b99d5a8acd200}}
\promptfield{Prompt tokens}{\mechex{10{,}060}}
\end{promptmeta}

\promptsection{Rendered Prompt Excerpt}
\textcolor{taskprompttemplate}{Audit independent qualitative propagation
traces.}

\textcolor{taskprompttemplate}{Restart from the referenced context's initial
domains for each case. At every displayed step, the before-set must equal the
target variable's current domain. A cited law supports a target cell exactly
when the other distinct variables can take cells from their current domains so
that the qualitative sum is satisfiable: all terms zero, or at least one
positive and one negative term.}

\begin{promptitems}
\item \textbf{Context \mechex{\code{X01}}:}
\mechex{Selected states include \code{conduction.REVERSE_HEAT_FLOW},
\code{mass1.HEATING}, and \code{mass2.COOLING};
\code{conduction.G} starts in \{positive\},
\code{conduction.port_a.Q_flow} in \{negative\}, and seven other displayed
domains in \{negative, zero, positive\}.}
\item \textbf{Displayed laws:}
\mechex{\code{Tsensor1.law_50ac08c818ee2f2e},
\code{Tsensor1.law_50ac08c818ee2f2e.d1},
\code{Tsensor2.law_50ac08c818ee2f2e},
\code{Tsensor2.law_50ac08c818ee2f2e.d1},
\code{conduction.law_63510044227e162d},
\placeholder{five additional laws}.}
\item \textbf{Case \mechex{\code{C001}}, context
\mechex{\code{X01}}:}
\mechex{Steps are
\code{[1,Tsensor1.d1__port__Q_flow,[negative,zero,positive],[zero],
Tsensor1.law_50ac08c818ee2f2e]},
\code{[2,Tsensor2.port.Q_flow,[negative,zero,positive],[zero],
Tsensor2.law_50ac08c818ee2f2e]},
\code{[3,Tsensor2.d1__port__Q_flow,[negative,zero,positive],[zero],
Tsensor2.law_50ac08c818ee2f2e.d1]},
\code{[4,conduction.Q_flow,[negative,zero,positive],[negative],
conduction.law_af234bf2c7beec25]},
\code{[5,conduction.port_b.Q_flow,[negative,zero,positive],[positive],
Tsensor1.law_50ac08c818ee2f2e]}, and
\code{[6,conduction.dT,[negative,zero,positive],[negative,zero],
conduction.law_63510044227e162d]}.}
\item \textbf{Additional task data:}
\mechex{\placeholder{remaining_contexts_and_cases}.}
\end{promptitems}

\textcolor{taskprompttemplate}{Multiply factor signs normally; use the
supplied cell-sign maps. A step is faithful iff the cited law contains the
target and its exact supported target-cell set equals the displayed after-set.
Apply every displayed after-set even after an unfaithful citation. Cases are
independent. The step fields are
\code{[index,target,before,after,cited_law_id]}.}

\promptsection{Required Output Format}
\begin{prompttemplateblock}
\{"answer":\{"unfaithful\_step\_count":0\}\}
\end{prompttemplateblock}
\end{promptbox}

\subsubsection{Answer extraction and normalization rules}

Extraction follows this ordered procedure.

\begin{enumerate}
\item Read the backend text response.
\item Parse the complete response.
\item If complete-response parsing fails and a matching fenced object is
present, parse that object. If the matched fence is malformed, extraction fails
immediately.
\item Only if no matching fence is present, parse the text spanning the first
opening brace through the last closing brace, if present.
\item If the parsed value is an object with an \texttt{answer} field, use that
field; otherwise use the parsed value itself as the family-specific payload.
\item Normalize the family-specific payload and compare its canonical form with
the simulator gold object when computing accuracy.
\end{enumerate}

Every family normalizer projects the parsed object onto the documented family
payload, so unrelated top-level fields are discarded. The expected answer is
used only to determine the requested keys or fixed output width, not to repair
or infer model-supplied values. The six canonicalization rules are:

\smallskip
\noindent\textbf{State consistency.}
Only requested scenario ids are retained. Missing scenarios and requested scenarios whose
values are not objects become zero consistent candidates, while unrequested
scenarios and fields are discarded. Candidate ids are stripped; invalid ids are
discarded, and ids matching \code{S[0-9]+} are deduplicated and sorted by
their numeric suffix. A count may be an integer, an integral number, or
a decimal digit string; an omitted count defaults to the number of retained
ids. Malformed or negative counts and count--id mismatches are rejected.

\smallskip
\noindent\textbf{Plausibility.}
The current plausibility labels payload must be exactly the requested
32-character width and contain only \code{P} or \code{I}, in query-id order.
No case folding, trimming, or label repair is applied.

\smallskip
\noindent\textbf{Necessity.}
The current necessity labels payload must be exactly the requested
four-character width and contain only \code{N} or \code{U}, in query-id order.
No case folding, trimming, or label repair is applied.

\smallskip
\noindent\textbf{Episode-graph transitions.}
Admissible component state transitions must be an array whose entries
contain exactly component, source state, and target state
with nonempty string values. Duplicate triples are rejected; accepted triples
are sorted lexicographically by those three fields before comparison.

\smallskip
\noindent\textbf{Functional recovery.}
Plan statuses must have exactly the requested width and contain only
\code{U}, \code{D}, or \code{R}, in displayed policy order. No case folding,
trimming, or status repair is applied.

\smallskip
\noindent\textbf{Trace faithfulness.}
Unfaithful step count must be an integer; booleans, numeric strings,
and nonintegral numbers are rejected. The accepted integer is compared directly
with the simulator count.

Rate limits wait and retry automatically. Other backend or transport failures
are persisted and may be targeted manually with the unchanged prompt; the final
attempt and its history are retained. Retries do not combine samples or repair
an invalid answer after the model has responded.

The outcome classes and reported metrics are defined in
Section~\ref{app:outcome-separated-metrics}.

\subsection{Per-Task Family Complexity Curves}
\label{app:task-family-curves}

The curves appear in Figure~2 of the main paper. To construct their B1--B4
analysis bins, unique retained items within each task family are ordered by
$(C(i),\text{mechanism id},\text{task id})$. Items with equal $C(i)$ form an
indivisible tie block. Each of the three cuts is placed at the tie-block
boundary nearest its cumulative equal-count target; an exact distance tie
selects the earlier boundary. This deterministic rule produces near-equal
quartiles while preserving every tie block.

\subsection{OpenModelica--Qualitative Consistency Check}
\label{app:validation-portfolio}

OpenModelica supplies a numerical execution path separate from the qualitative
solver. For fixed Modelica instances, its persisted traces are projected onto
the qualitative component-state vocabulary and compared with qualitative
results only after the numerical scenarios have been frozen. The comparison
reads no benchmark tasks or gold solutions and has no path to admission,
generation, labeling, or scoring. Here, ``independent'' means that the numerical
and qualitative results follow separate execution paths and that the comparison
is task- and gold-independent; it does not mean clean-room implementation. The
comparison is limited to behavior observed in the finite runs.

\paragraph{Pinned execution environment.}
The deterministic regression suite pins OpenModelica 1.26.9 and the DASSL
solver in \code{openmodelica/openmodelica:v1.26.9-minimal}. The validation
configuration records the exact run method, tolerance, time window, output
grid, initial-value overrides, and artifact hashes. Settings not explicitly
overridden remain OpenModelica's DASSL defaults.

Each fixed Modelica instance is compiled once. The bounded suite includes a
nominal run and, where applicable, source-cycle, one-at-a-time
initial-coordinate perturbation, and numerical-robustness runs. It does not
construct a Cartesian product of initial conditions or perturb mechanism
parameters. Source-cycle runs are checked only after projection to selected
qualitative properties; isolated numerical samples are not mapped to benchmark
episode nodes.

\paragraph{Frozen comparison protocol.}
The check freezes the OpenModelica observations before qualitative comparison.
\code{extract} records each distinct composite-state vector and observed
component transition together with its run and timestamp. A numerical value
within $10^{-9}$ of a point landmark retains the snapped cell and the cell
containing the raw value. \code{predict} checks each frozen state for exact MDD
nonemptiness and each transition for a directed path in the declared transition
graph. \code{score} joins the frozen observations and qualitative results by
stable scenario identifier.

A state is a direct agreement when its crisp relation is nonempty. When that
relation is empty, the observation is boundary-compatible exactly when every
frozen raw-side boundary-candidate group contains at least one alternative with
a nonempty exact relation. It is a contradiction when the crisp relation and
at least one complete raw-side candidate group are both exactly empty. A
transition agrees when its declared directed path exists. A crisp or
boundary-candidate solve that exceeds its explicit work limit before either
disposition is established is inconclusive.

\par\noindent
\begin{minipage}{\columnwidth}
\centering
\small
\tabcolsep=4pt
\resizebox{\columnwidth}{!}{%
\begin{tabular}{lrrrrr}
\toprule
Scenario type & Observed & Direct & Boundary & Contradict & Inconcl. \\
\midrule
Composite states & 186 & 184 & 2 & 0 & 0 \\
Transition shapes & 155 & 155 & 0 & 0 & 0 \\
\midrule
Total & 341 & 339 & 2 & 0 & 0 \\
\bottomrule
\end{tabular}
}
\captionof{table}{OpenModelica--qualitative consistency check for 341 distinct observed
scenarios from 72 pinned OpenModelica experiments across all 18 catalog
mechanisms. The frozen rule classifies 339 as direct agreements and two as
boundary-compatible; none is contradictory or inconclusive.}
\label{tab:behavioral-concordance}
\end{minipage}\par

\begin{table*}[!t]
\centering
\small
\tabcolsep=2pt
\renewcommand{\arraystretch}{0.9}
\begin{tabularx}{\textwidth}{>{\raggedright\arraybackslash}p{2.8cm} >{\raggedright\arraybackslash}p{2.15cm} >{\raggedright\arraybackslash}p{2.65cm} >{\raggedright\arraybackslash}X >{\raggedright\arraybackslash}p{2.1cm}}
\toprule
Model & Scale & Architecture & Documented profile & Capability class \\
\midrule
GPT-5.5 \cite{gpt55modelcard} & Frontier (undisclosed) & Undisclosed GPT architecture & Reasoning, coding, tool-heavy agents, and long-running tasks; 1.05M context & Agentic-capable reasoning LLM \\
o3 \cite{openai2025o3} & (undisclosed) & Undisclosed reasoning architecture & Multi-step reasoning across text, code, and images; function calling & Agentic-capable reasoning LLM \\
DeepSeek-R1 \cite{deepseekai2025r1} & 671B total / 37B active & Sparse MoE transformer & Reasoning model based on DeepSeek-V3-Base with cold-start data and reinforcement learning & Reasoning LLM \\
gpt-oss-20b \cite{openai2025gptoss} & 21B total / 3.6B active & Sparse MoE transformer & Reasoning post-training and interleaved web, Python, and developer-tool use & Agentic-capable reasoning LLM \\
Qwen3-Coder-30B \cite{qwenteam2025qwen3coder} & 30.5B total / 3.3B active & Sparse MoE transformer & Code-specialized, repository-scale, agentic coding and browser use & Agentic-capable coding LLM \\
GPT-4.1 \cite{openai2025gpt41} & (undisclosed) & Undisclosed GPT architecture & Instruction following, coding, tool use, long context, and powering agents & Agentic-capable LLM \\
Llama-3.3-70B \cite{meta2024llama33} & 70B & Dense autoregressive transformer with GQA & General-purpose multilingual model with documented tool-use evaluation & Tool-capable LLM \\
Qwen3-30B \cite{qwenteam2025qwen3} & 30.5B total / 3.3B active & Sparse MoE transformer & Multilingual reasoning with thinking modes and external-tool integration & Agentic-capable reasoning LLM \\
Gemma-3-12B \cite{google2025gemma3} & 12B & Dense decoder-only transformer & General-purpose multilingual, code and math exposure; 12T-token pretraining & General LLM \\
\bottomrule
\end{tabularx}
\caption{Model set spanning scale, architecture, reasoning, and documented
agentic capability, ordered by descending accuracy. Capability classes summarize
the cited first-party documentation.}
\label{tab:baseline-matrix}
\medskip

\begin{minipage}{\textwidth}
\centering
\small
\tabcolsep=4pt
\begin{tabularx}{\textwidth}{
  >{\raggedright\arraybackslash}p{2.5cm}
  >{\raggedright\arraybackslash}X
  >{\raggedleft\arraybackslash}p{2.0cm}
  >{\raggedright\arraybackslash}p{2.15cm}}
\toprule
Display label & Exact evaluator profile & Effective output-token limit &
Reasoning-effort request \\
\midrule
GPT-5.5 & \code{gpt-5.5} & 20{,}000 & low \\
o3 & \code{o3} & 20{,}000 & low \\
DeepSeek-R1 & \code{DeepSeek-R1} & 20{,}000 & not sent \\
gpt-oss-20b & \code{openai/gpt-oss-20b} & 20{,}000 & low \\
Qwen3-Coder-30B & \code{qwen/qwen3-coder-30b-a3b-instruct} &
20{,}000 & not sent \\
GPT-4.1 & \code{gpt-4.1} & 20{,}000 & not sent \\
Llama-3.3-70B & \code{Llama-3.3-70B-Instruct} & 8{,}192 & not sent \\
Qwen3-30B & \code{qwen/qwen3-30b-a3b} & 16{,}384 & low \\
Gemma-3-12B & \code{google/gemma-3-12b-it} & 16{,}384 & not sent \\
\bottomrule
\end{tabularx}
\captionof{table}{Exact evaluator profiles and negotiated output and reasoning controls.
``Not sent'' means that the endpoint profile does not expose a configurable
reasoning-effort parameter; it does not classify the model as non-reasoning.}
\label{tab:evaluator-request-controls}
\end{minipage}
\end{table*}

\paragraph{Results and boundary scenarios.}
The two boundary-compatible states occur in
\code{mechanics_branched_three_mass} at 0.001\,s and
\code{mechanics_damper} at 0.735 or 0.757\,s. In both scenarios, the $10^{-9}$
landmark tolerance maps small positive velocities to \code{REST}, while the
scaled damper forces remain positive. The raw equations
$f_1=5v_{\mathrm{leaf1}}$, $f_2=8v_{\mathrm{leaf2}}$, and $f=25v$ remain
satisfied, and exact MDD checks accept the corresponding
\code{MOVING_POSITIVE} states.
These landmark effects yield no physical
contradiction.

\paragraph{Coverage boundary.}
Simulation-based validation is limited to behavior observed in the 72 finite
runs and therefore cannot validate existential claims for which no supporting
behavior was observed. The qualitative abstraction itself follows established
formulations in the qualitative-reasoning literature
\cite{forbus1984,kuipers1984,kuipers1986,klenk2014modelica}.

\subsection{Model Inventory}
\label{app:model-inventory}

Table~\ref{tab:baseline-matrix} summarizes the evaluated model inventory.
All evaluators use the direct-LLM protocol in
Section~\ref{app:prompt-evaluator}; display labels map to the persisted profiles
in Table~\ref{tab:evaluator-request-controls}. We call an LLM
\emph{agentic-capable} only when first-party sources document or evaluate
closed-loop, multi-step tool use: selecting actions toward a goal,
updating decisions from observations, and continuing, revising, or stopping.
Reasoning, long context, or isolated function calling alone is insufficient.
\emph{Tool-capable} denotes structured tool calling without such evidence;
\emph{general} denotes no documented tool-use capability; and \emph{reasoning}
is an independent, provider-documented modifier. An \emph{agentic system}
additionally supplies orchestration, tools, state, policies, and stopping
conditions \cite{openai2025agentguide}. Because our evaluations use no such
scaffold, these classes describe documented model capabilities, not the
evaluated configurations.

\subsection{Released Mechanism Inventory}
\label{app:worked-mechanisms}

This inventory reports provenance, task-family support, and retained counts for
the 18 released mechanisms.

\begin{table*}[!t]
\begin{minipage}{\textwidth}
\centering
\tabcolsep=2pt
\renewcommand{\arraystretch}{1.0}
\begin{tabularx}{\textwidth}{>{\raggedright\arraybackslash}p{5.05cm} >{\raggedright\arraybackslash}p{2.75cm} >{\raggedright\arraybackslash}X >{\raggedleft\arraybackslash}p{1.3cm}}
\toprule
Catalog mechanism id & Provenance & Supported task families & Retained \\
\midrule
\code{electrical_electrothermal_resistor_parallel} & Repository composition & SC, FR, TF & 48 \\
\code{electrical_resistor} & MSL 4.1.0 & SC, P, N, T, FR, TF & 68 \\
\code{electrical_electrothermal_resistor_ladder} & Repository composition & SC, P, N, FR, TF & 69 \\
\code{mechanics_branched_three_mass} & Repository composition & SC, FR, TF & 44 \\
\code{mechanics_compare_braking_force} & MSL 4.1.0 & SC, P, N, T, FR, TF & 96 \\
\code{mechanics_compare_braking_torque} & MSL 4.1.0 & SC, P, N, T, FR, TF & 96 \\
\code{mechanics_first} & MSL 4.1.0 & SC, FR, TF & 44 \\
\code{mechanics_grounded_geared_dual_inertia} & Repository composition & SC, P, N, T, FR, TF & 64 \\
\code{mechanics_sign_convention} & MSL 4.1.0 & SC, N, T, FR & 44 \\
\code{mechanics_accelerate} & MSL 4.1.0 & P, N, T, FR, TF & 68 \\
\code{mechanics_damper} & MSL 4.1.0 & SC, P, N, T, FR & 53 \\
\code{mechanics_first_grounded} & MSL 4.1.0 & SC, FR, TF & 44 \\
\code{mechanics_initial_conditions} & MSL 4.1.0 & SC, N, FR, TF & 56 \\
\code{mechanics_oscillator} & MSL 4.1.0 & SC, N, T, FR, TF & 52 \\
\code{mechanics_why_arrows} & MSL 4.1.0 & SC, P, N, T, FR & 60 \\
\code{thermal_branched_star} & Repository composition & SC, P, N, T, FR & 60 \\
\code{thermal_two_masses} & MSL 4.1.0 & SC, P, N, FR, TF & 65 \\
\code{thermal_three_node_chain} & Repository composition & SC, P, N, T, FR, TF & 89 \\
\bottomrule
\end{tabularx}
\captionof{table}{Exact released catalog inventory.  SC denotes state consistency, P plausibility, N necessity,
T episode-graph transitions, FR functional recovery, and TF trace faithfulness.
``Supported'' means that the release retained at least one task for the
family--mechanism cell.}
\label{tab:worked-mechanisms}
\end{minipage}
\end{table*}

\onecolumn
\clearpage
\raggedbottom
\begin{multicols}{2}
\bibliography{references}
\end{multicols}
\end{document}